\pdfoutput=1

\documentclass[11pt]{article}

\usepackage[]{EMNLP2022}

\usepackage{times}
\usepackage{latexsym}

\usepackage[T1]{fontenc}

\usepackage[utf8]{inputenc}

\usepackage{microtype}

\usepackage{inconsolata}

\usepackage{comment}
\usepackage{amsmath,amsfonts,amssymb,amsthm,mathtools}
\usepackage{graphicx}
\usepackage{xcolor}
\usepackage{color,soul}
\usepackage{booktabs}
\usepackage{mathtools}
\usepackage{bbm}
\usepackage{linguex}
\usepackage{subfigure}
\usepackage{array}
\usepackage{float}
\usepackage{tikz}
\usepackage[algoruled,ruled,vlined,noend]{algorithm2e}
\SetAlFnt{\normalsize}
\SetAlCapFnt{\normalsize}
\SetAlCapNameFnt{\normalsize}

\definecolor{colorpath1}{HTML}{999999}
\definecolor{colorpath2}{HTML}{1764AB}
\definecolor{colorpath3}{HTML}{E27878}
\definecolor{yellowpaper}{HTML}{D6B656}
\definecolor{bluepaper}{HTML}{6C8EBF}
\definecolor{redpaper}{HTML}{E19090}
\definecolor{green_smooth}{HTML}{ADC977}

\usepackage{bm}

\renewcommand{\vec}[1]{\bm{#1}}

\newcommand{\vx}{\vec{x}}
\newcommand{\vxtil}{\widetilde{\vec{x}}}

\newcommand{\ve}{\vec{e}}

\newcommand{\redmath}[2]{%
  \tikz[baseline]
  \node[fill=redpaper, rounded corners=3pt, anchor=base]
  (#1)
  {\ensuremath{#2}};}
 \newcommand{\greenmath}[2]{%
  \tikz[baseline]
  \node[fill=green_smooth, rounded corners=3pt, anchor=base]
  (#1)
  {\ensuremath{#2}};}

\usepackage[T1]{fontenc}

\usepackage[utf8]{inputenc}

\usepackage{microtype}
\newcommand\norm[1]{\left\lVert#1\right\rVert}
\newcommand{\eos}{\texttt{<\text{/s}>}}

\definecolor{color0}{HTML}{FFFFFF}
\definecolor{color1}{HTML}{E3EEF9}
\definecolor{color2}{HTML}{D0E1F2}
\definecolor{color3}{HTML}{B7D4EA}
\definecolor{color4}{HTML}{94C4DF}
\definecolor{color5}{HTML}{6AAED6}
\definecolor{color6}{HTML}{4A98C9}
\definecolor{color7}{HTML}{2E7EBC}
\definecolor{color8}{HTML}{1764AB}

\usepackage{nameref}
\usepackage{cleveref}

\crefformat{section}{\S#2#1#3}
\crefformat{subsection}{\S#2#1#3}
\crefformat{subsubsection}{\S#2#1#3}
\crefformat{paragraph}{\P#2#1#3}
\crefformat{subparagraph}{\P#2#1#3}
\crefmultiformat{section}{\S#2#1#3}{ and~\S#2#1#3}{, \S#2#1#3}{, and~\S#2#1#3}
\crefmultiformat{subsection}{\S#2#1#3}{ and~\S#2#1#3}{, \S#2#1#3}{, and~\S#2#1#3}
\crefmultiformat{subsubsection}{\S#2#1#3}{ and~\S#2#1#3}{, \S#2#1#3}{, and~\S#2#1#3}
\crefmultiformat{paragraph}{\P\P#2#1#3}{ and~#2#1#3}{, #2#1#3}{, and~#2#1#3}
\crefmultiformat{subparagraph}{\P\P#2#1#3}{ and~#2#1#3}{, #2#1#3}{, and~#2#1#3}
\crefrangeformat{section}{\mbox{\S\S#3#1#4--#5#2#6}}
\crefrangeformat{subsection}{\mbox{\S\S#3#1#4--#5#2#6}}
\crefrangeformat{subsubsection}{\mbox{\S\S#3#1#4--#5#2#6}}
\crefrangeformat{paragraph}{\mbox{\P\P#3#1#4--#5#2#6}}
\crefrangeformat{subparagraph}{\mbox{\P\P#3#1#4--#5#2#6}}
\crefname{part}{Part}{Parts}
\Crefname{part}{Part}{Parts}
\crefname{chapter}{ch.}{ch.}
\Crefname{chapter}{Ch.}{Ch.}
\crefname{footnote}{fn.}{fn.}
\Crefname{footnote}{Fn.}{Fn.}
\crefname{figure}{figure}{figures}
\crefname{subfigure}{figure}{figures}
\Crefname{subfigure}{Figure}{Figures}
\crefname{appsec}{appendix}{appendices}
\Crefname{appsec}{Appendix}{Appendices}
\crefname{algocf}{algorithm}{algorithms}
\Crefname{algocf}{Algorithm}{Algorithms}

\crefname{ExNo}{example}{examples}
\Crefname{ExNo}{Example}{Examples}
\crefname{SubExNo}{example}{examples}
\Crefname{SubExNo}{Example}{Examples}
\crefname{SubSubExNo}{example}{examples}
\Crefname{SubSubExNo}{Example}{Examples}
\crefformat{ExNo}{(#2#1#3)}
\crefformat{SubExNo}{(#2#1#3)}
\crefformat{SubSubExNo}{(#2#1#3)}
\crefrangeformat{ExNo}{\mbox{(#3#1#4--#5#2#6)}}
\crefrangeformat{SubExNo}{\mbox{(#3#1#4--#5#2#6)}}
\crefrangeformat{SubSubExNo}{\mbox{(#3#1#4--#5#2#6)}}
\crefmultiformat{ExNo}{(#2#1#3}{, #2#1#3)}{, #2#1#3}{, #2#1#3)}
\crefmultiformat{SubExNo}{(#2#1#3}{, #2#1#3)}{, #2#1#3}{, #2#1#3)}
\crefmultiformat{SubSubExNo}{(#2#1#3}{, #2#1#3)}{, #2#1#3}{, #2#1#3)}
\crefrangemultiformat{ExNo}{(#3#1#4--#5#2#6}{, #3#1#4--#5#2#6)}{, #3#1#4--#5#2#6}{, #3#1#4--#5#2#6)}
\crefrangemultiformat{SubExNo}{(#3#1#4--#5#2#6}{, #3#1#4--#5#2#6)}{, #3#1#4--#5#2#6}{, #3#1#4--#5#2#6)}
\crefrangemultiformat{SubSubExNo}{(#3#1#4--#5#2#6}{, #3#1#4--#5#2#6)}{, #3#1#4--#5#2#6}{, #3#1#4--#5#2#6)}

\title{Towards Opening the Black Box of Neural Machine Translation: \newline Source and Target Interpretations of the Transformer}

\author{Javier Ferrando$^1$, Gerard I. Gállego$^1$, Belen Alastruey$^1$,\\
    {\bf Carlos Escolano$^1$, Marta R. Costa-jussà$^2$}\\
         $^1$TALP Research Center, Universitat Politècnica de Catalunya \\
         $^2$Meta AI \\
         \texttt{\{javier.ferrando.monsonis,gerard.ion.gallego,} \\ \texttt{belen.alastruey,carlos.escolano\}@upc.edu}\\
         \texttt{costajussa@meta.com}
         }

\begin{document}
\maketitle
\begin{abstract}
In Neural Machine Translation (NMT), each token prediction is conditioned on the source sentence and the target prefix (what has been previously translated at a decoding step). However, previous work on interpretability in NMT has mainly focused solely on source sentence tokens' attributions. Therefore, we lack a full understanding of the influences of every input token (source sentence and target prefix) in the model predictions. In this work, we propose an interpretability method that tracks input tokens' attributions for both contexts. Our method, which can be extended to any encoder-decoder Transformer-based model, allows us to better comprehend the inner workings of current NMT models. We apply the proposed method to both bilingual and multilingual Transformers and present insights into their behaviour.
\end{abstract}

\section{Introduction}
Transformers \cite{NIPS2017_3f5ee243} have become the state-of-the-art architecture for natural language processing (NLP) tasks \cite{devlin-etal-2019-bert,T5_raffel,NEURIPS2020_1457c0d6}. With its success, the NLP community has experienced an urge to understand the decision process of the model predictions  \cite{jain-wallace-2019-attention,serrano-smith-2019-attention}.

In Neural Machine Translation (NMT), attempts to interpret Transformer-based predictions have mainly focused on analyzing the attention mechanism \cite{raganato-tiedemann-2018-analysis, voita-etal-2018-context}. A large number of works in this line have investigated the capabilities of the cross-attention to perform source-target alignment \cite{kobayashi-etal-2020-attention,Zenkel_2019,chen-etal-2020-accurate}, compared with human annotations. Gradient-based \cite{ding-etal-2019-saliency} and occlusion-based methods \cite{li-etal-2019-word} have also been evaluated against human word alignments. The former computes gradients with respect to the input token embeddings to measure how much a change in the input changes the output, the latter generates input attributions by measuring the change in the predicted probability after deleting specific tokens. However, there is a tension between finding a faithful explanation and observing human-like alignments, since one does not imply the other \cite{ferrando-costa-jussa-2021-attention-weights}.

\begin{figure}[!t]
\begin{center}
\includegraphics[width=0.49\textwidth]{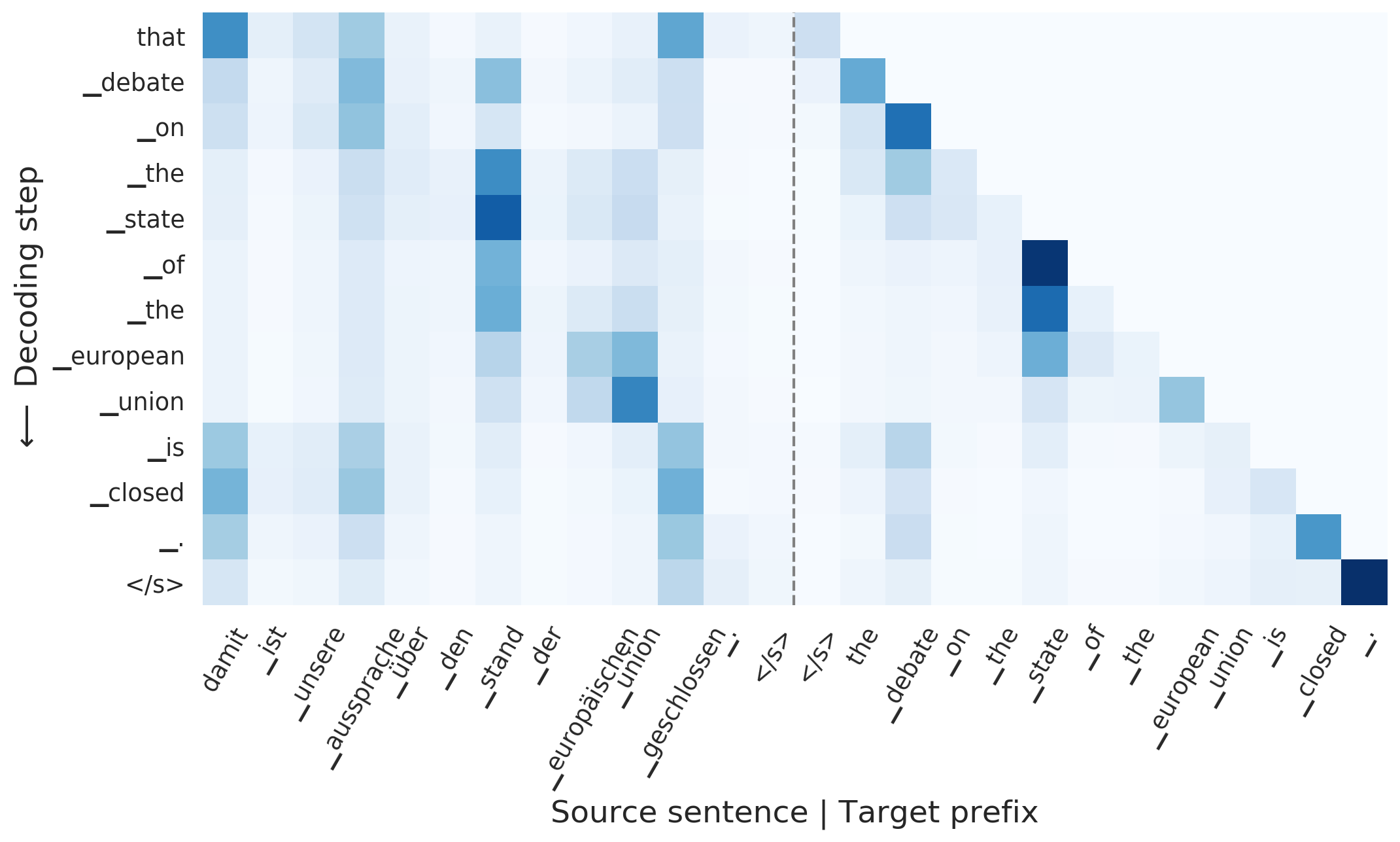}
\caption{ALTI+ results for a De-En translation example. We obtain source sentence and target prefix (columns) interpretations for every predicted token (row).}
\label{fig:rollout_l6}
\end{center}
\end{figure}

The decoding process of NMT systems consists of generating tokens in the target vocabulary based on the information provided by the source sequence and the previously generated tokens (target prefix). However, most of the work on interpretability of NMT models only analyses source tokens. Recently, \citet{voita-etal-2021-analyzing} proposed using Layer Relevance Propagation (LRP) \cite{LRP_bach} to analyze the source and target contributions to the model prediction, and later analyzed its behaviour during training \citep{voita-etal-2021-language}. Nonetheless, they apply their method to obtain global explanations, as an average over the entire dataset, not to get input attributions of a single prediction. Gradient-based methods have also been extended to the target prefix \cite{ferrando-costa-jussa-2021-attention-weights}, although they do not quantify the relative contribution of source and target inputs.

Concurrently, encoder-based Transformers, such as BERT \cite{devlin-etal-2019-bert} and RoBERTa \cite{DBLP:journals/corr/abs-1907-11692}, have been analysed with attention rollout \cite{abnar-zuidema-2020-quantifying}, which models the information flow in the model with a Directed Acyclic Graph, where nodes are token representations and edges, attention weights. In the computer vision literature, \citet{Chefer_2021_CVPR,Chefer_2021_ICCV} combined this method with gradient information. Recently, \citet{ferrando2022measuring} have presented ALTI (Aggregation of Layer-wise Tokens Attributions), which applies the attention rollout method by substituting attention weights with refined token-to-token interactions. In this work, we present the first application of a rollout-based method to sequence to sequence Transformers. Our key contributions are\footnote{Code available at \url{https://github.com/mt-upc/transformer-contributions-nmt}.}:
\begin{itemize}
    \item We propose a method that measures the contributions of each input token (source and target prefix) to the encoder-decoder Transformer predictions;
    \item We show how contextual information is mixed across the encoder of NMT models, with the model keeping up to 47\% of token identity;
    \item We evaluate the role of residual connections in the cross-attention, and show that attention to uninformative source tokens (EOS and final punctuation mark) is used to let information flow from the target prefix;
    \item We analyze the role of both input contexts in low and high-resource scenarios, and show the model behaviour under hallucinations.
\end{itemize}

\section{Background}

In this section, we provide the background to understand our proposed method by briefly explaining the encoder-decoder Transformer-based model in the context of NMT \cite{NIPS2017_3f5ee243} and the Aggregation of Layer-wise Token-to-token Interactions (ALTI) method \cite{ferrando2022measuring}.

\subsection{Encoder-Decoder Transformer}
\begin{figure}[!t]
\begin{center}\includegraphics[width=0.5\textwidth]{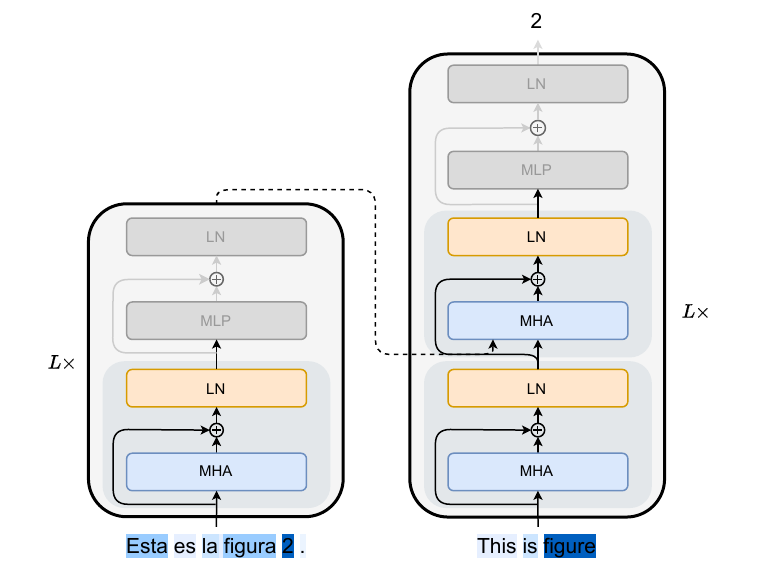}
\caption{Encoder-Decoder Transformer.}
\label{fig:enc_dec}
\end{center}
\end{figure}

Given a source sequence of tokens $\mathbf{x} = (x_1, \ldots, x_{J})$, and a target sequence $\mathbf{y} = (y_1, \ldots, y_{T})$, an NMT system models the conditional probability:
\begin{equation}
P(\mathbf{y}|\mathbf{x}) = \prod_{t=1}^{T} P(y_t|\mathbf{y}_{<t},\mathbf{x})
\end{equation}
where $\mathbf{y}_{<t} = (y_0, \ldots, y_{t-1})$ represents the prefix of $y_{t}$, with $x_{J} = y_{0} = \eos$ used as a special token to mark the beginning and end of sentence. The Transformer is composed by a stack of encoder and decoder layers (\Cref{fig:enc_dec}). The encoder generates a contextualized sequence of representations $\mathbf{e} = (\ve_1, \ldots, \ve_J)$ of the source sentence. The decoder, at each time step $t$, uses both the encoder outputs ($\mathbf{e}$) and the target prefix ($\mathbf{y}_{<t}$) to compute a probability distribution over the target vocabulary, from which a prediction is sampled.

\paragraph{Multi-head attention.} The Transformer core building block, the multi-head attention mechanism (MHA) is in charge of combining contextual information in the hidden representations. Consider here $\mathbf{x} = (\vx_1, \ldots, \vx_J)$ as the sequence of token representations\footnote{We consider $\vx_i$ as a column vector.} of dimension $d$ entering layer $l$, and $\tilde{\mathbf{x}} = (\vxtil_1, \ldots, \vxtil_{J})$ the output layer representations. Each of the $H$ heads inside MHA computes vectors of dimension $d_h = d/H$:
\begin{equation}\label{eq:head_output}
\bm z^{h}_i = \sum_{j=1}^J  \mathbf{\alpha}_{i,j}^{h} \mathbf{W}_V^{h}\bm x_{j}
\end{equation}
with $\mathbf{\alpha}_{i,j}^{h}$ referring to the attention weight where token $i$ attends token $j$, and $\mathbf{W}_V^{h} \in \mathbb{R}^{d_h \times d}$ to a learned weight matrix\footnote{The bias vector associated with $\mathbf{W}_V^{h}$ is omitted for the sake of simplicity.}.

The output of MHA for the i-th token ($\text{MHA}_i$) is calculated by concatenating each $\bm z^{h}_i$ and projecting the joint vector through $\mathbf{W}_O \in \mathbb{R}^{d \times d}$. This is equivalent to a sum over heads where each $\bm z^{h}_i$ is projected through the partitioned weight matrix $\mathbf{W}_O^{h} \in \mathbb{R}^{d \times d_h}$ and adding the bias $\bm b_{O} \in \mathbb{R}^d$:
\begin{equation}\label{eq:mha_output1}
\begin{aligned}
\text{MHA}_i(\mathbf{x}) &= \mathbf{W}_O\;\text{Concat}(\bm z^{1}_i, \ldots, \bm z^{H}_i) + \bm b_O\\
&= \sum^H_{h=1} \mathbf{W}_O^{h} \bm z^{h}_i + \bm b_O
\end{aligned}
\end{equation}
\paragraph{Layer normalization.}
Finally, a layer normalization (LN) is applied over the sum of the residual vector $\bm x_i$ and the output of the multi-head attention module, giving as output $\widetilde{\bm x}_i$:
\begin{equation}\label{eq:post_layer_output}
\widetilde{\bm x}_i = \text{LN}(\text{MHA}_i(\mathbf{x}) + \bm x_{i})
\end{equation}

Merging \Cref{eq:mha_output1,eq:head_output,eq:post_layer_output}, we get:
\begin{equation*}
\widetilde{\bm x}_i = \text{LN}\Bigg(\sum_{j=1}^J \sum^H_{h=1} \mathbf{W}_O^{h} \mathbf{\alpha}_{i,j}^{h} \mathbf{W}_V^{h}\bm x_{j} + \bm b_O + \bm x_{i}\Bigg)
\end{equation*}
Considering $F_{i}(\bm x_j) = \sum^H_{h=1} \mathbf{W}_O^{h} \mathbf{\alpha}_{i,j}^{h} \mathbf{W}_V^{h}\bm x_{j}$, we can formulate the previous equation as:
\begin{equation}\label{eq:mha_output2}
\widetilde{\bm x}_i = \text{LN}\Bigg(\sum_{j=1}^J F_{i}(\bm x_j) + \bm b_O + \bm x_{i}\Bigg)
\end{equation}
\subsection{Aggregation of Layer-wise Token-to-token Interactions (ALTI)}\label{sec:alti}
\begin{figure}[!t]
\begin{center}\includegraphics[width=0.49\textwidth]{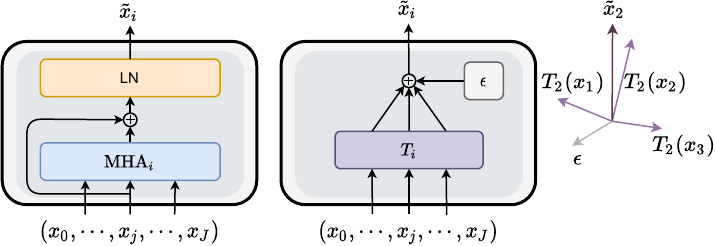}
\caption{The self-attention block (left) at each position $i$ can be decomposed as a summation of transformed input vectors (right). The closest vector ($T_{2}(x_2)$) \textit{contributes} the most to $\widetilde{\bm x}_{2}$.}
\label{fig:contribs_explanation}
\end{center}
\end{figure}

The layer normalization operation over a sum of vectors $\text{LN}(\sum_j \bm u_j)$, as in \Cref{eq:mha_output2}, can be reformulated as $\sum_j L(\bm u_j) + \bm \beta$, where $L: \mathbb{R}^d \mapsto \mathbb{R}^d$ (see \Cref{apx:ln}). This allows us to express \Cref{eq:mha_output2} \cite{kobayashi-etal-2021-incorporating} as an interpretable expression of the layer input representations (\Cref{fig:contribs_explanation}):
\begin{equation}\label{eq:post_layer_transformed_vectors}
\widetilde{\bm x}_i = \sum_{j=1}^J T_i(\bm x_j) + \bm \epsilon
\end{equation}

where $\bm \epsilon$ contains bias terms (see \Cref{apx:full_derivation} for full derivation) and $T_i$ transforms the layer input vectors:
\begin{equation}\label{eq:transformed_vectors}
  T_i(\bm x_j)=\left\{
  \begin{array}{@{}ll@{}}
    L(F_{i}(\bm x_{j})) & \mbox{if}~ j \neq i \\
    L(F_{i}(\bm x_{j}) + \bm x_i) & \mbox{if}~ j=i
  \end{array}\right.
\end{equation}

with the residual connection $\bm x_{i}$ only considered in the transformed vector $T_{i}(\bm x_{j=i})$. \citet{ferrando2022measuring} propose to use the Manhattan distance between the output vector and the transformed vector as a measure of the impact of $\bm x_{j}$ on $\widetilde{\bm x}_i$:
\begin{equation}\label{eq:distance}
d_{i,j} = \norm{\widetilde{\bm x}_i- T_i(\bm x_j)}_1
\end{equation}
By taking $-d_{i,j}$, larger distances reflect lower (more negative) influence. Then, distances are normalized $\in [0,1]$ to obtain the \textit{contribution of token representation j to token representation i}\footnote{We use the term `contribution' to refer to influences between token representations. `Relevance' is used to allude to the influence of input tokens to model predictions.}:
\begin{equation}\label{eq:contributions}
c_{\widetilde{\bm x}_{i} \gets \bm x_j} = \frac{\text{max}(0,-d_{i,j} + \norm{\widetilde{\bm x}_i}_1)}{\sum_{k=1}^{J} \text{max}(0, -d_{i,k} + \norm{\widetilde{\bm x}_i}_1)}
\end{equation}
giving  the matrix of layer-wise contributions $\mathbf{C}_{\widetilde{\bm x}_{i} \gets \mathbf{x}} \in \mathbb{R}^{J \times J}$, where each row contains the contribution, or influence, of each $\bm x_j$ in $\widetilde{\bm x}_i$.

ALTI method \cite{ferrando2022measuring} follows the Transformer's modeling approach proposed by \citet{abnar-zuidema-2020-quantifying}, where the information flow in the model is simplified as a Directed Acyclic Graph, where nodes are token representations, and edges represent the influence of each input layer token $\bm x_j$ in the output token $\widetilde{\bm x}_i$. ALTI proposes using token contributions $\mathbf{C}$ instead of raw attention weights $\alpha$. The amount of information flowing from one node to another in different layers is computed by summing over the different paths connecting both nodes, where each path is the result of the multiplication of every edge in the path. This is computed by the matrix multiplication of the layer-wise contributions, giving the full encoder contribution matrix:
\begin{equation}\label{eq:rollout}
\mathbf{C}^{\text{enc}}_{\mathbf{e} \gets \mathbf{x}} = \mathbf{C}_{\mathbf{e} \gets \mathbf{x}}^{L} \cdot \mathbf{C}_{\widetilde{\mathbf{x}} \gets \mathbf{x}}^{L-1} \cdot \; \cdots \; \cdot \mathbf{C}_{\widetilde{\mathbf{x}} \gets \mathbf{x}}^{1}
\end{equation}

We refer to $\mathbf{C}_{\mathbf{e} \gets \mathbf{x}}^{L}$ as the contributions in the last layer of the encoder, where output vectors are $\mathbf{e}$. 

\section{ALTI for the Encoder-Decoder Transformer (ALTI+)}
The attention rollout and ALTI methods work for encoder-based Transformers. However, in the encoder-decoder Transformer, the cross-attention hinders its integration. In this section, we present ALTI+, which is the adaptation of ALTI method to the encoder-decoder Transformer.
\subsection{Decoder Layer Decomposition}\label{sec:decoder_layer_decomposition}
\begin{figure}[t]
\begin{center}
\includegraphics[width=0.48\textwidth]{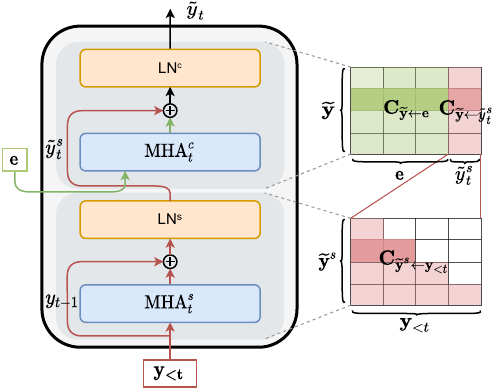}
\caption{Self-attention and cross-attention modules in a decoder layer together with its contribution matrices.\protect\footnotemark In green, it's shown the information coming from the encoder (source), and in red, the information from the decoder (target prefix). Highlighted is shown contributions at a single time step $t$.}
\label{fig:cross_diagram}
\end{center}
\end{figure}

We decompose the self-attention and cross-attention of a decoder layer into interpretable expressions (\Cref{eq:post_layer_transformed_vectors}), from which we can get the degree of interaction between input and output token representations (\Cref{eq:contributions}). Consider $\mathbf{y}_{<t} = (\bm y_0, \ldots, \bm y_{j}, \ldots, \bm y_{t-1})$ the set of vector representations of the target prefix tokens as input of a decoder layer, and $\widetilde{\bm y}_t$ the layer output (\Cref{fig:cross_diagram}).

\footnotetext{We omit the MLP and its LN of the decoder layer.}
\paragraph{Decoder self-attention.}\label{sec:dec_selfattn_decomposition}
The layer normalization in the decoder self-attention ($\text{LN}^s$) is applied over the sum of the multi-head attention output and the residual $\bm y_{t-1}$. The self-attention block\footnote{We refer as `block' to the multi-head attention, residual, and layer normalization.} can be written as:
\begin{equation}\label{eq:cross_self_output}
\begin{aligned}
\widetilde{\bm y}^{s}_t &= \text{LN}^{s}(\text{MHA}^s_t(\mathbf{y}_{<t}) + \bm y_{t-1})\\
&= \text{LN}^{s}\left(\sum_{j=0}^{t-1} F^{s}_t(\bm y_{j}) + \bm b_O + \bm y_{t-1}\right)
\end{aligned}
\end{equation}
where $F^s_{t}$ considers $\alpha, \mathbf{W}^h_V$ and $\mathbf{W}^h_O$ of the decoder self-attention. Analogous to \Cref{eq:transformed_vectors} we can obtain the transformed vectors of $\bm y_j$:
\begin{equation*}
  T^s_t(\bm y_j)=\left\{
  \begin{array}{@{}ll@{}}
    L^s (F^{s}_{i}(\bm y_{j})) & \mbox{if}~ j \neq t-1 \\
    L^s (F^{s}_{i}(\bm y_{j}) + \bm y_{t-1}) & \mbox{if}~ j=t-1
  \end{array}\right.
\end{equation*}
Following \Cref{eq:distance,eq:contributions} we get the decoder self-attention contributions $\mathbf{C}_{\widetilde{\mathbf{y}}^{s}\gets \mathbf{y}_{<t}} \in \mathbb{R}^{T \times T}$ reflecting the strength of the interaction between $\mathbf{y}_{<t}$ and $\widetilde{\bm y}^s_{t}$.

\paragraph{Decoder cross-attention.}\label{sec:cross_attn_decomposition}
The output of the cross-attention block at time step $t$ can be decomposed as:
\begin{equation}\label{eq:cross_output}
\begin{aligned}
\widetilde{\bm y}_t &= \text{LN}^{c}(\text{MHA}^c_t(\greenmath{}{\mathbf{e}}) + \redmath{}{\widetilde{\bm y}^{s}_{t}})\\
&= \text{LN}^{c}\left(\sum_{j=1}^J F^{c}_t(\greenmath{}{\bm e_{j}}) + \bm b_O + \redmath{}{\widetilde{\bm y}^{s}_{t}}\right)
\end{aligned}
\end{equation}
where $\widetilde{\bm y}^{s}_{t}$, the residual connection, is the output of the self-attention block, and $\mathbf{e}$ the encoder outputs. We can obtain the transformed vectors of the encoder outputs $\bm e_j$ and the residual connection $\widetilde{\bm y}^{s}_t$:
\begin{equation}
\begin{aligned}
T^{c}_{t}(\bm e_j) &= L^c(F^c_t(\bm e_{j}))\\
T^{c}_{t}(\bm y^{s}_{t}) &=   L^c(\widetilde{\bm y}^{s}_{t})
\end{aligned}
\end{equation}
Following \Cref{eq:distance}, we can compute the Manhattan distance between the transformed vectors and $\widetilde{\bm y}_{t}$ and get the contributions $[\mathbf{C}_{\widetilde{\mathbf{y}}\gets\mathbf{e}};\mathbf{C}_{\widetilde{\mathbf{y}}\gets\widetilde{\bm{y}}^{s}_{t}}]$, with $\mathbf{C}_{\widetilde{\mathbf{y}}\gets\mathbf{e}} \in \mathbb{R}^{T \times J}$ and $\mathbf{C}_{\widetilde{\mathbf{y}}\gets\widetilde{\bm{y}}^{s}_{t}} \in \mathbb{R}^{T \times 1}$. 

\begin{figure}[H]
\begin{center}
\includegraphics[width=0.34\textwidth]{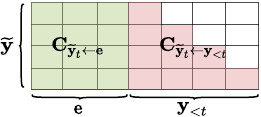}
\caption{Full decoder layer contributions.} 
\label{fig:contrib_matrix_decoder}
\end{center}
\end{figure}

The cross-attention residual $\widetilde{\bm y}^s_t$ contribution to $\widetilde{\bm y}_{t}$ reflects the total influence of the self-attention inputs $\mathbf{y}_{<t}$ to the decoder layer output $\widetilde{\bm y}_t$. Thus, we can get the full \textit{decoder layer contribution matrix} $[\mathbf{C}_{\widetilde{\mathbf{y}}\gets\mathbf{e}};\mathbf{C}_{\widetilde{\mathbf{y}}\gets\mathbf{y}_{<t}}]$ (\Cref{fig:contrib_matrix_decoder}) by substituting the residual contributions ($\mathbf{C}_{\widetilde{\mathbf{y}}\gets\widetilde{\bm{y}}^{s}_{t}}$) with the self-attention contributions ($\mathbf{C}_{\widetilde{\mathbf{y}}\gets\widetilde{\bm{y}}^{s}_{t}}$), and weighting every row of $\mathbf{C}_{\widetilde{\mathbf{y}}\gets\mathbf{y}_{<t}}$ by the corresponding value of the residual contribution of each time step.

\begin{figure*}[!t]
\centering
\begin{minipage}{.45\textwidth}
  \centering
  \includegraphics[width=0.91\linewidth]{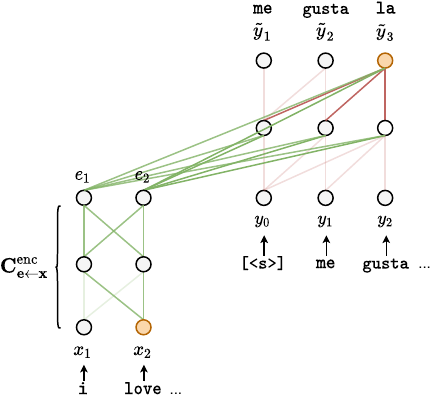}
  \captionof{figure}{Source input attributions $\mathbf{R}^{\text{model}}_{\widetilde{\mathbf{y}}_t \gets \mathbf{x}}$.}
  \label{fig:alti_source}
\end{minipage}%
\hspace{1.5cm}
\begin{minipage}{.45\textwidth}
  \centering
  \includegraphics[width=0.9\linewidth]{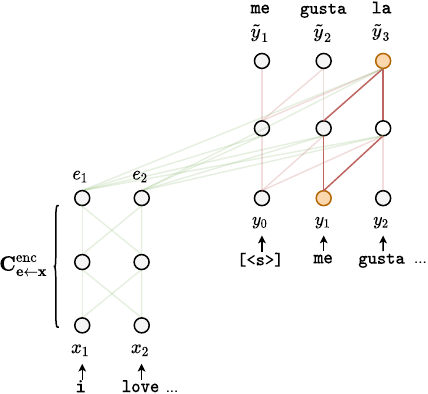}
  \captionof{figure}{Target prefix input attributions $\mathbf{R}^{\text{model}}_{\widetilde{\mathbf{y}}_t \gets \mathbf{\mathbf{y}_{<t}}}$.}
  \label{fig:alti_target}
\end{minipage}
\end{figure*}

\subsection{Aggregating Contributions Through the Encoder-Decoder Transformer}
In order to get \textit{input token attributions}, we apply the same principle as attention rollout method. As described in \Cref{sec:alti}, ALTI builds a graph where nodes are token representations and edges represent the contributions between tokens in each layer. The amount of information flowing from one node to another in different layers is computed by summing over the different paths connecting both nodes, where each path is the result of the multiplication of every edge in the path (\Cref{fig:alti_source,fig:alti_target}).

\paragraph{ALTI+ source tokens relevance.}
\begin{algorithm}[h!]
     \SetAlgoNoLine
     \SetNoFillComment
     \KwIn{$\mathbf{C}^{\text{enc}}_{\mathbf{e} \gets \mathbf{x}}$ -- encoder contributions\newline
            $\mathbf{C}^{l}_{\widetilde{\mathbf{y}}_t \gets \mathbf{e}}$ -- contributions decoder layers\newline
            $L$ -- number of layers}
    \KwOut{$\mathbf{R}^{\text{model}}_{\widetilde{\mathbf{y}}_t \gets \mathbf{x}}$ -- source input relevancies}
     \SetKwInOut{KwIn}{Input}
    \SetKwInOut{KwOut}{Output}
     \For{l $\gets$ [1,2...L]}{
     $\mathbf{C}^{*l}_{\widetilde{\mathbf{y}}_t \gets \mathbf{x}} = \mathbf{C}^{l}_{\widetilde{\mathbf{y}}_t \gets \mathbf{e}} \cdot \mathbf{C}^{\text{enc}}_{\mathbf{e} \gets \mathbf{x}}$ \\
     }
     $\mathbf{R}^{1}_{\widetilde{\mathbf{y}}_t \gets \mathbf{x}} = \mathbf{C}^{*1}_{\widetilde{\mathbf{y}}_t \gets \mathbf{x}}$\\
     \For{l $\gets$ [2,3...L]}{
     $\mathbf{R}^{l}_{\widetilde{\mathbf{y}}_t \gets \mathbf{x}} = \mathbf{C}^{l}_{\widetilde{\mathbf{y}}_t \gets \mathbf{y}_{<t}} \cdot \mathbf{R}^{l-1}_{\widetilde{\mathbf{y}}_t \gets \mathbf{x}} + \mathbf{C}^{*l}_{\widetilde{\mathbf{y}}_t \gets \mathbf{x}} $\\
     }
     $\mathbf{R}^{\text{model}}_{\widetilde{\mathbf{y}}_t \gets \mathbf{x}}$ = $\mathbf{R}^{L}_{\widetilde{\mathbf{y}}_t \gets \mathbf{x}}$\\
     \Return $\mathbf{R}^{\text{model}}_{\widetilde{\mathbf{y}}_t \gets \mathbf{x}}$
     
 \caption{ALTI+ source relevance.}
 \captionsetup{font=small, skip=0pt}
 \label{algo:ALTI_source}
 \end{algorithm}
\Cref{algo:ALTI_source} shows the process to obtain source sentence tokens relevance for the model prediction $R^{\text{model}}_{\widetilde{\mathbf{y}}_t \gets \mathbf{x}}$ (\Cref{fig:alti_source}).
We first update the cross-attention contribution matrices (to $\mathbf{C}^{*l}_{\widetilde{\mathbf{y}}_t \gets \mathbf{x}}$) by multiplying each of them with the contributions of the entire encoder $\mathbf{C}^{\text{enc}}_{\mathbf{e} \gets \mathbf{x}}$ to account for all the paths in the encoder and cross-attentions. We then iteratively aggregate edges from paths of the target prefix contributions $\mathbf{C}^{l}_{\widetilde{\mathbf{y}}_t \gets \mathbf{y}_{<t}}$.

\paragraph{ALTI+ target prefix tokens relevance.}
Target prefix input attributions (\Cref{fig:alti_target}) are computed by multiplying $\mathbf{C}_{\widetilde{\mathbf{y}} \gets \mathbf{y}_{<t}}$ in each layer:
\begin{equation}\label{eq:alti_target}
\mathbf{R}^{\text{model}}_{\widetilde{\mathbf{y}}_t \gets \mathbf{\mathbf{y}_{<t}}} = \mathbf{C}^{L}_{\widetilde{\mathbf{y}}\gets\mathbf{y}_{<t}} \cdot \mathbf{C}^{L-1}_{\widetilde{\mathbf{y}} \gets \mathbf{y}_{<t}} \cdot \; \cdots \; \cdot \mathbf{C}^{1}_{\widetilde{\mathbf{y}} \gets \mathbf{y}_{<t}}
\end{equation}

\section{Experimental Setup}
We analyze input token attributions in both bilingual and multilingual Machine Translation models. For the bilingual setting, we train a 6-layer Transformer model for the German-English (De-En) translation task. We use Europarl v7 corpus\footnote{\url{http://www.statmt.org/europarl/v7}} and follow \citet{Zenkel_2019} and \citet{ding-etal-2019-saliency} data setup\footnote{\url{https://github.com/lilt/alignment-scripts/tree/master/preprocess}}. We use byte-pair encoding (BPE) \cite{sennrich-etal-2016-neural} with 10k merge operations. For the multilingual model, we use M2M Transformer \cite{m2m_100}, a many-to-many multilingual translation model that can translate directly between any pair of 100 languages. We use \textsc{Fairseq} \cite{ott-etal-2019-fairseq} implementations, and the provided checkpoint for the M2M model (418M). We perform the quantitative analysis in 1000 sentences of the test set of IWSLT’14 German-English dataset. For the analysis in \Cref{sec:multilingual} we use \textsc{Flores-101} \cite{goyal-etal-2022-flores} devtest split.


\begin{figure*}[!t]
    \centering
    \subfigure[Attention weights]{\includegraphics[width=0.265\textwidth]{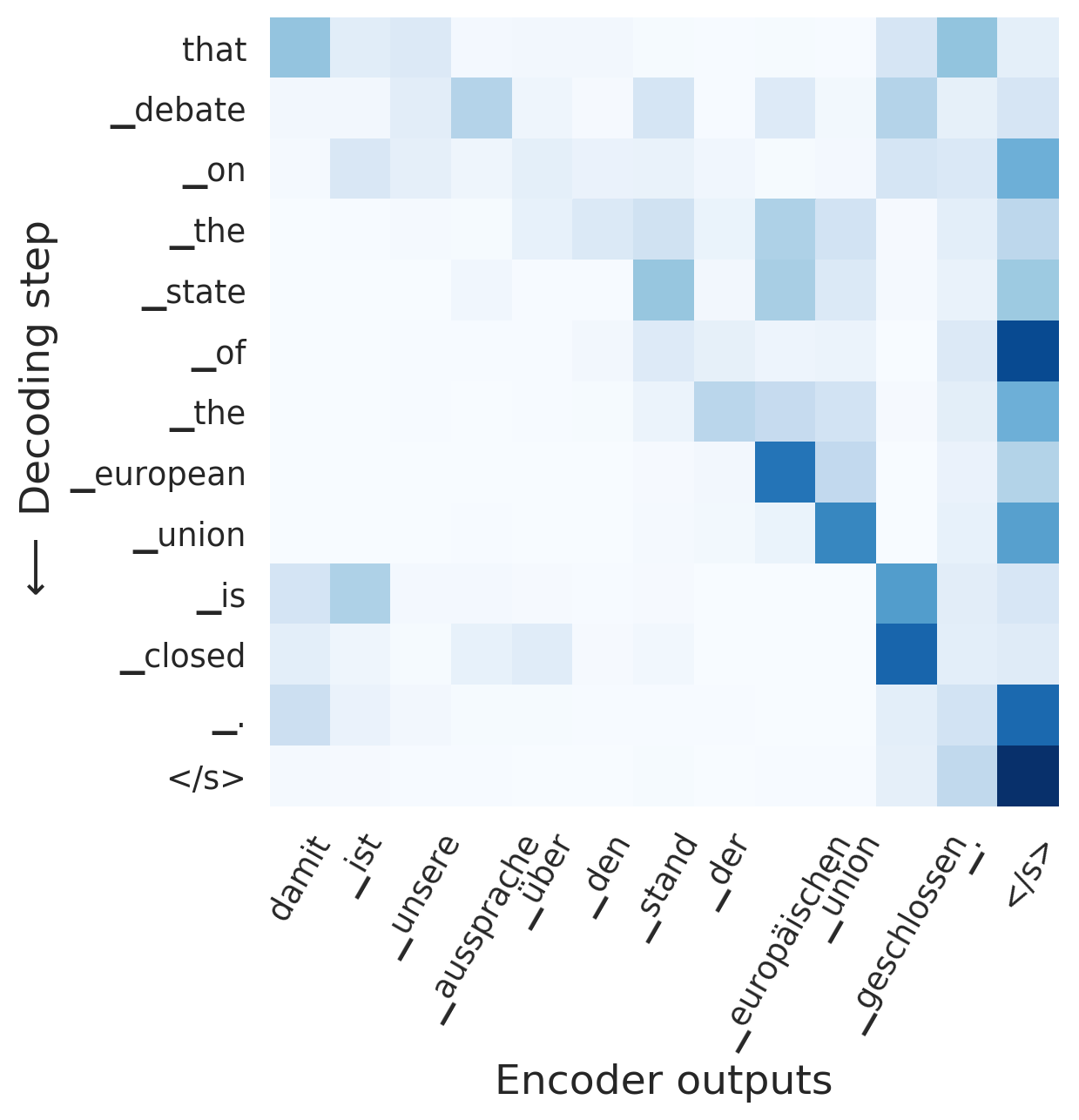}} 
    \subfigure[Contributions]{\includegraphics[width=0.274\textwidth]{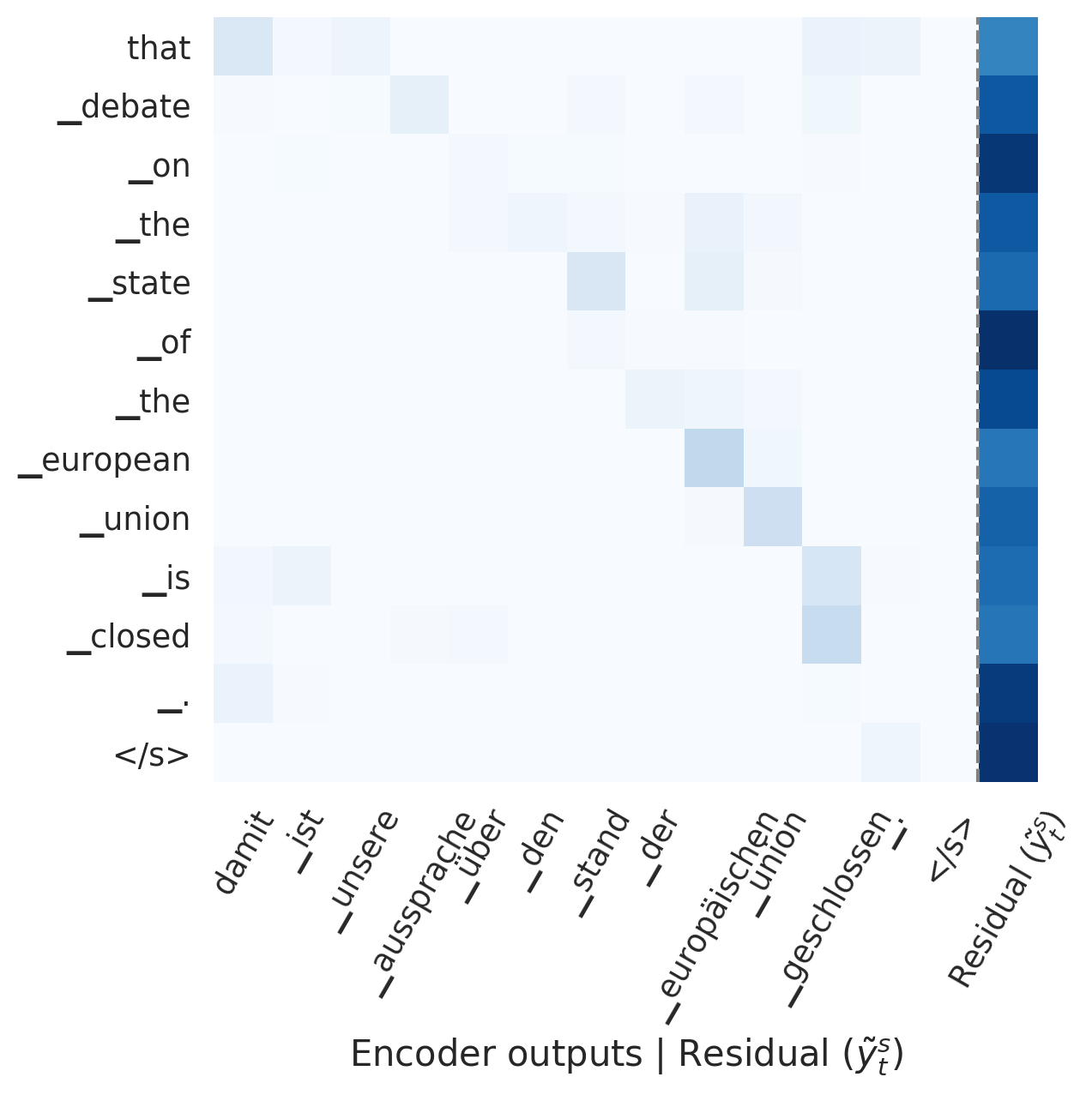}} 
    \subfigure[Decoder layer contributions]{\includegraphics[width=0.450\textwidth]{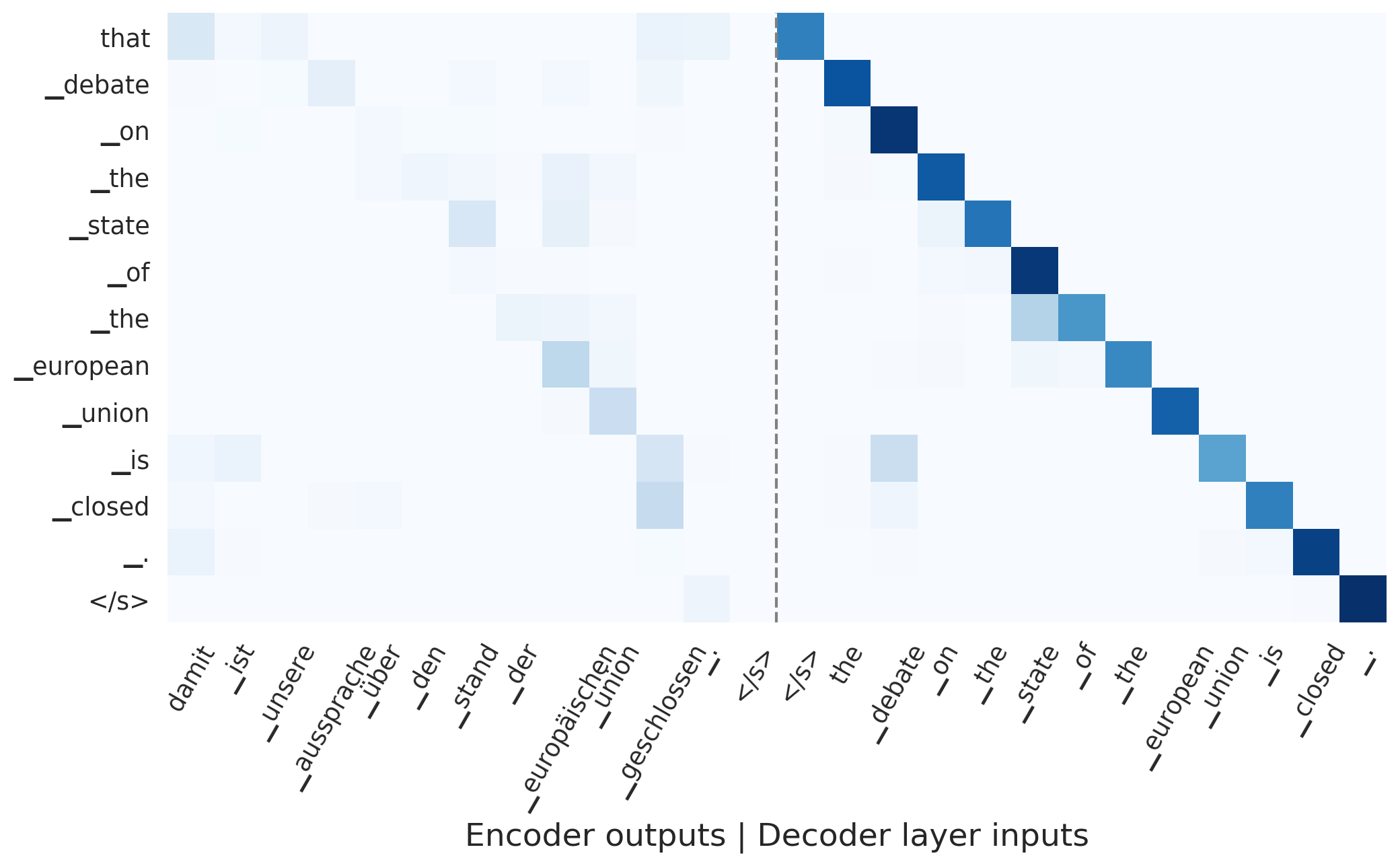}}
    \caption{(a) Cross-attention weights. (b) Cross-attention contributions $[\mathbf{C}_{\widetilde{\mathbf{y}}\gets\mathbf{e}};\mathbf{C}_{\widetilde{\mathbf{y}}\gets\widetilde{\bm{y}}^{s}_{t}}]$ of the encoder outputs $\mathbf{e}$ and residual $\widetilde{\bm y}_{t}^{s}$ to the decoder layer output, as described in \Cref{sec:cross_attn_decomposition}. (c) Total decoder layer contributions $[\mathbf{C}_{\widetilde{\mathbf{y}}\gets\mathbf{e}};\mathbf{C}_{\widetilde{\mathbf{y}}\gets\mathbf{y}_{<t}}]$ with the self-attention contributions included.}
    \label{fig:cross_attention_example}
\end{figure*}

\section{Analysis}
In this section, we perform a set of experiments to measure the quality of the obtained contributions, and unveil different aspects of bilingual and multilingual NMT models.

\begin{figure}[!t]
\begin{center}
\includegraphics[width=0.48\textwidth]{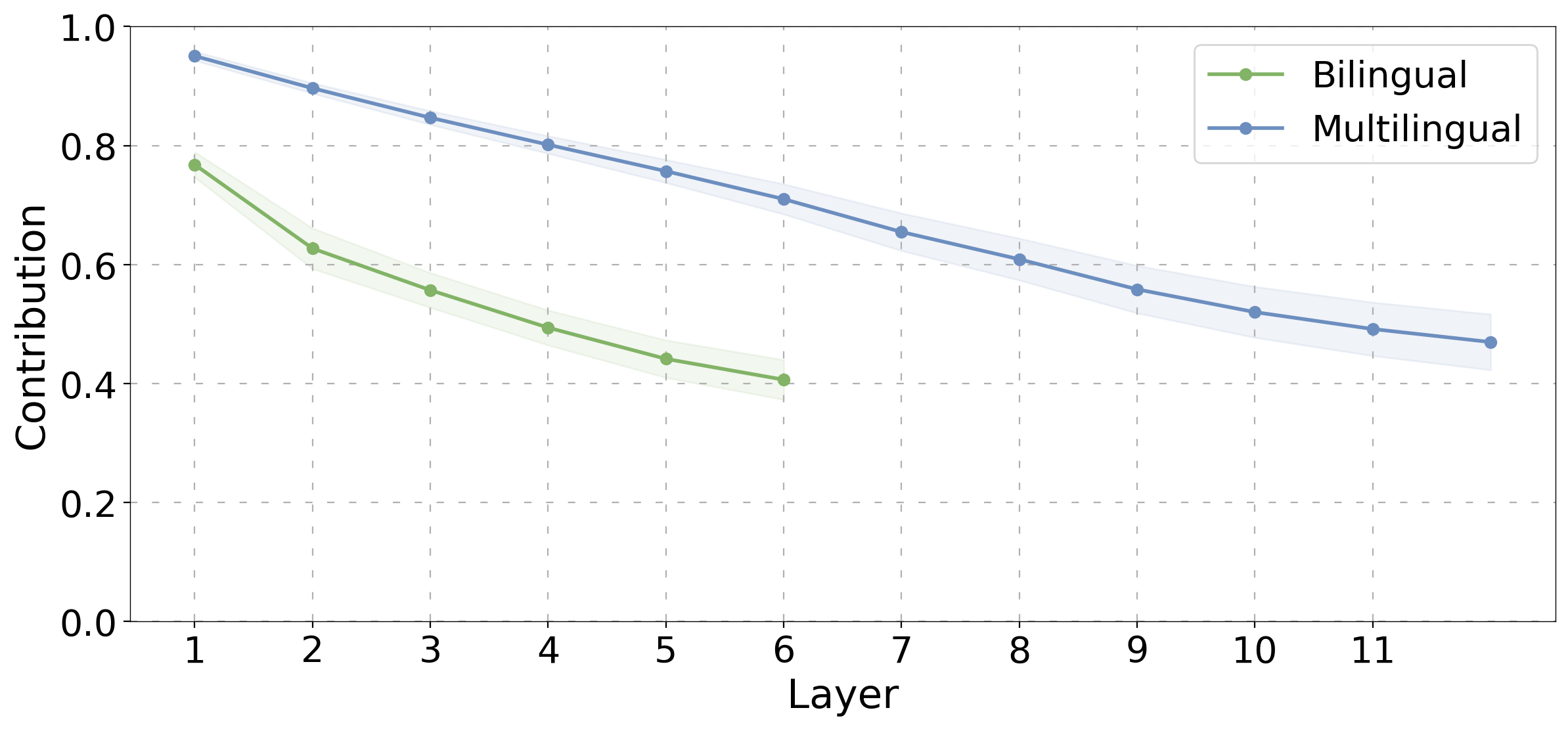}
\caption{Contribution of the source input token to the encoder output representation at the same position. We show mean and SD for each layer of the bilingual and multilingual models.}
\label{fig:bil_multi_enc_mix}
\end{center}
\end{figure}

\subsection{Information Mix in the Encoder}
Information from input source tokens gets mixed throughout the encoder. Intermediate layer representations acquire contextual information from other tokens in the sentence due to the self-attention mechanism. \citet{Brunner2020On} analyze, for an encoder-based model, the contribution of input source tokens to its intermediate layer representations. They conclude that input source tokens contribute little (around $10\%$ on average) to its corresponding last layer representation (encoder output). However, by training a linear classifier and, with nearest neighbor lookup based on the cosine distance, they are able to recover input token identity $93\%$ of the times. We apply ALTI method (\Cref{eq:rollout}) across the Transformer encoder and analyze the input relevance of source tokens to intermediate encoder representations (\Cref{fig:bil_multi_enc_mix}). Our results in the bilingual and multilingual models show that, indeed, input tokens highly contribute to their associated layer representations. In the last layer, 41\% of the input contribution comes from the input token at the same position. The multilingual model is able to retain above 47\% despite its 12 layers. The curves of both models in \Cref{fig:bil_multi_enc_mix} closely match the results obtained by \citet{voita-etal-2019-bottom} relying on the mutual information between the input tokens and tokens representations across layers.

\begin{table}[!t]
\centering
\resizebox{0.35\textwidth}{!}{%
\begin{tabular}{lc}
\toprule
\multicolumn{1}{c}{\textbf{Method}}  & \textbf{AER ($\downarrow$)}    \\ \midrule
Attention weights   & $47.7 \pm 1.7$                  \\
Vector-Norms & $41.4 \pm  1.4$                  \\
Vector-Norms + LN + Res & $42.5 \pm 0.8$                  \\
Our contributions $\mathbf{C}_{\widetilde{\mathbf{y}}\gets\mathbf{e}}$                         & $38.8 \pm 1.3$                  \\
\bottomrule

\end{tabular}
}
\caption{AER of the cross-attention contributions in the 5th layer of the bilingual model. We show mean and SD for models trained on five different seeds.}
\label{tab:aer_results}
\end{table}

\subsection{Alignment in Cross-attention}
In order to evaluate the quality of the proposed cross-attention contributions (\Cref{sec:decoder_layer_decomposition}), we measure Alignment Error Rate (AER) against human-annotated alignments.
As found out by \citet{garg-etal-2019-jointly}, the penultimate layer of Transformers tends to focus on learning the source-target alignment of words. Therefore, we analyze the cross-attention contributions $\mathbf{C}_{\widetilde{\mathbf{y}}\gets\mathbf{e}}$ extracted from the 5th layer from the bilingual 6-layer model. We use gold alignments from \citet{Vilar2006AERDW}, containing 508 sentence pairs. For comparison, we compute the AER of the raw attention weights and previous methods based on vector norms. \textit{Vector-Norms} \cite{kobayashi-etal-2020-attention} compute $\norm{F}_2$ from \Cref{eq:mha_output2}, and \textit{Vector-Norms + LN + Res} \cite{kobayashi-etal-2021-incorporating} $\norm{T}_2$ from \Cref{eq:post_layer_transformed_vectors}. As shown in \Cref{tab:aer_results}, our method for estimating layer-wise contributions obtain the lowest AER, outperforming similar previous methods by at least 2.6 points on average. As can be observed in \Cref{fig:cross_attention_example}, attention weights fail at showing alignments, with the $\eos$ token concentrating large attention weights. Our method is able to filter this noise, showing almost no contribution from \eos. In \Cref{sec:role_eos}, we analyze this phenomenon and try to find an explanation for it.

\begin{figure}[!t]
\begin{center}
\includegraphics[width=0.49\textwidth]{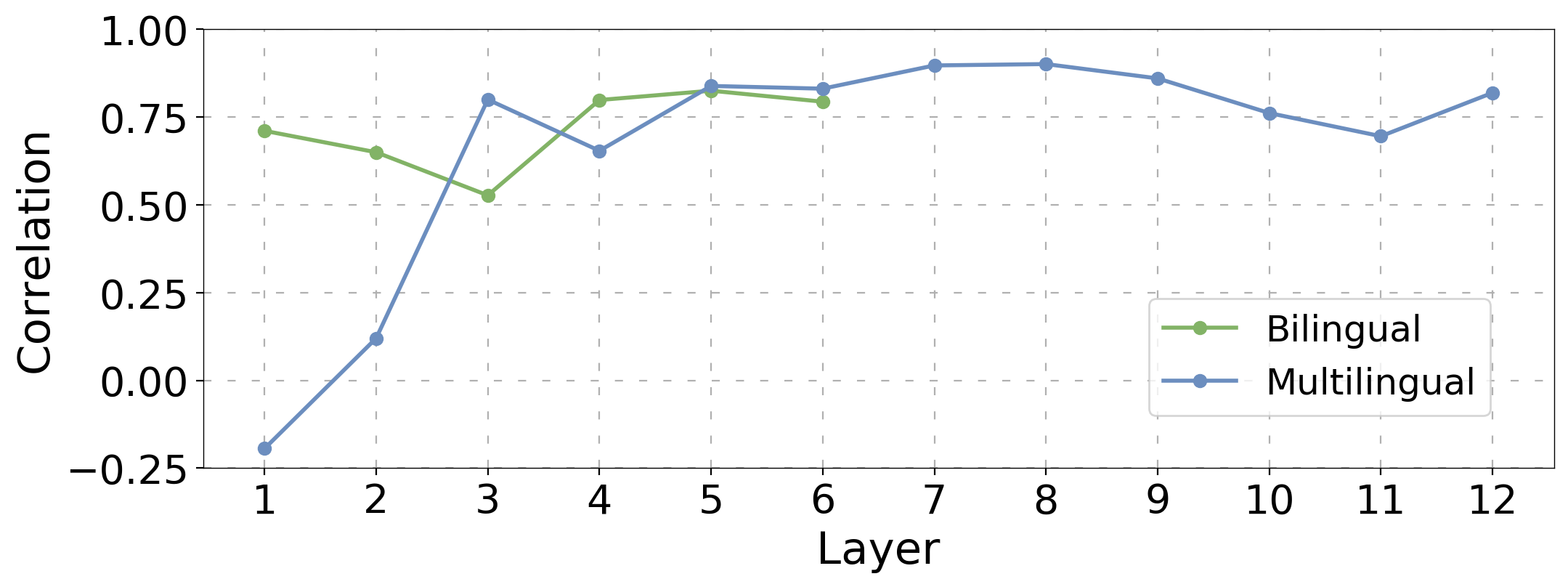}
\caption{Pearson's r correlation between attention weight values given to EOS token ($\eos$) and the contribution of the residual in the cross attention.}
\label{fig:corr_bil}
\end{center}
\end{figure}

\subsection{The Role of the End-of-Sentence Token}
\label{sec:role_eos}
It has been hypothesized that attention given to special tokens is used by the model as a `no-op' \cite{clark-etal-2019-bert}. \citet{ferrando-costa-jussa-2021-attention-weights} analyze attention weights of the cross-attention to source finalizing tokens (final punctuation mark and $\eos$), and find the value vectors (see \Cref{sec:values_norms}) associated with these tokens to be almost zero norm. Additionally, they find that attention weights to source finalizing tokens tend to increase when predicting tokens that heavily rely on the target prefix, such as postpositions, particles, or closing subwords. The proposed cross-attention decomposition in \Cref{sec:cross_attn_decomposition} allows us to analyze both the contributions of source tokens, and the residual connection (\Cref{fig:cross_attention_example} (b)). We measure the Pearson correlation between attention weights to $\eos$ token and the contribution of the residual connection in the cross-attention. We can see in \Cref{fig:corr_bil} that there is a high correlation in almost every layer, especially in the last layers. This demonstrates that finalizing tokens are used to \textit{skip source attention}, since the higher their attention score, the more information is flowing from the decoder (in the residual) coming from the target prefix.





\begin{figure}[!t]
\begin{center}
\includegraphics[width=0.48\textwidth]{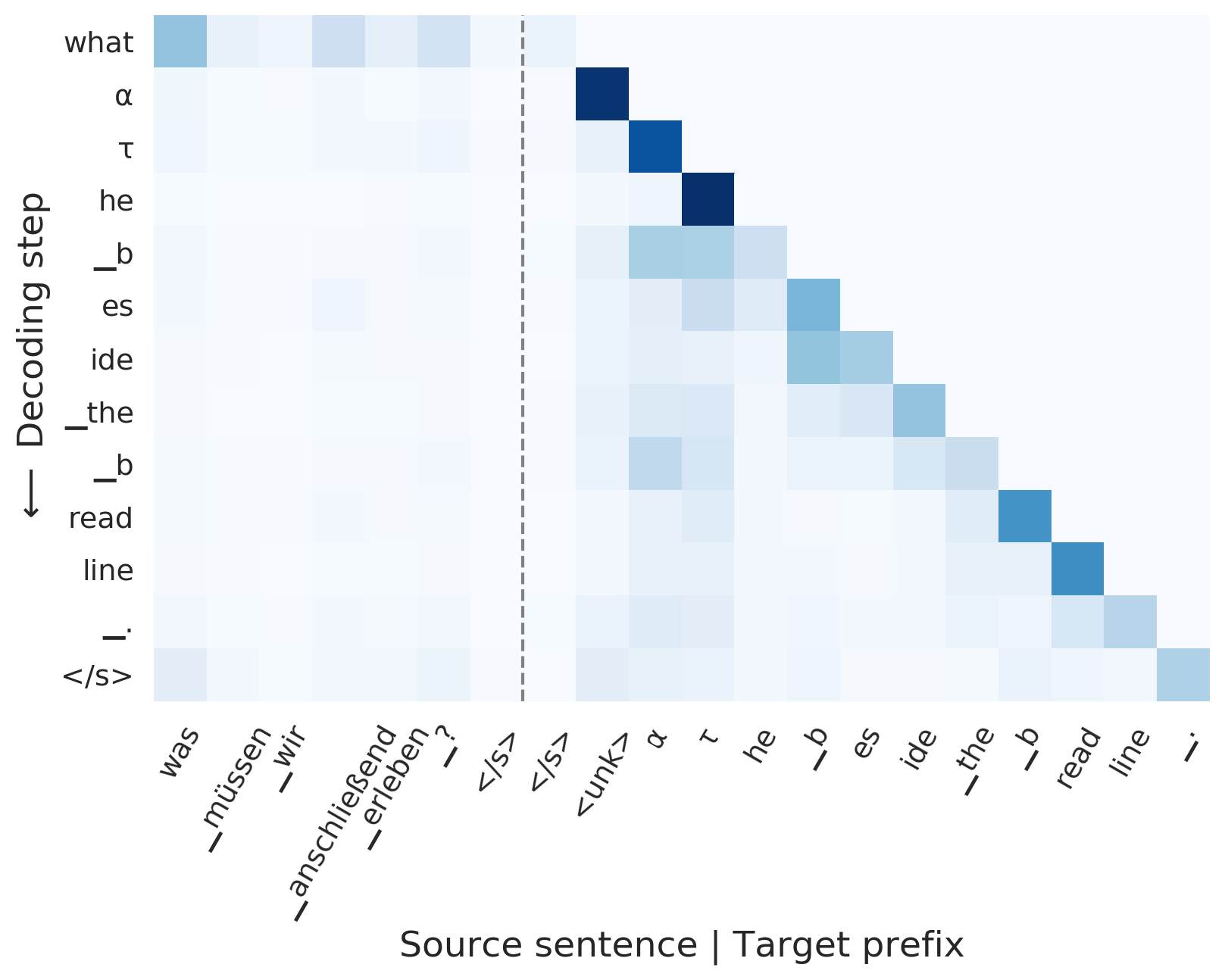}
\caption{ALTI+ results for a hallucination after induced perturbation in the bilingual model.}
\label{fig:hallucination_example}
\end{center}
\end{figure}

\begin{figure}[!t]
\begin{center}
\includegraphics[width=0.37\textwidth]{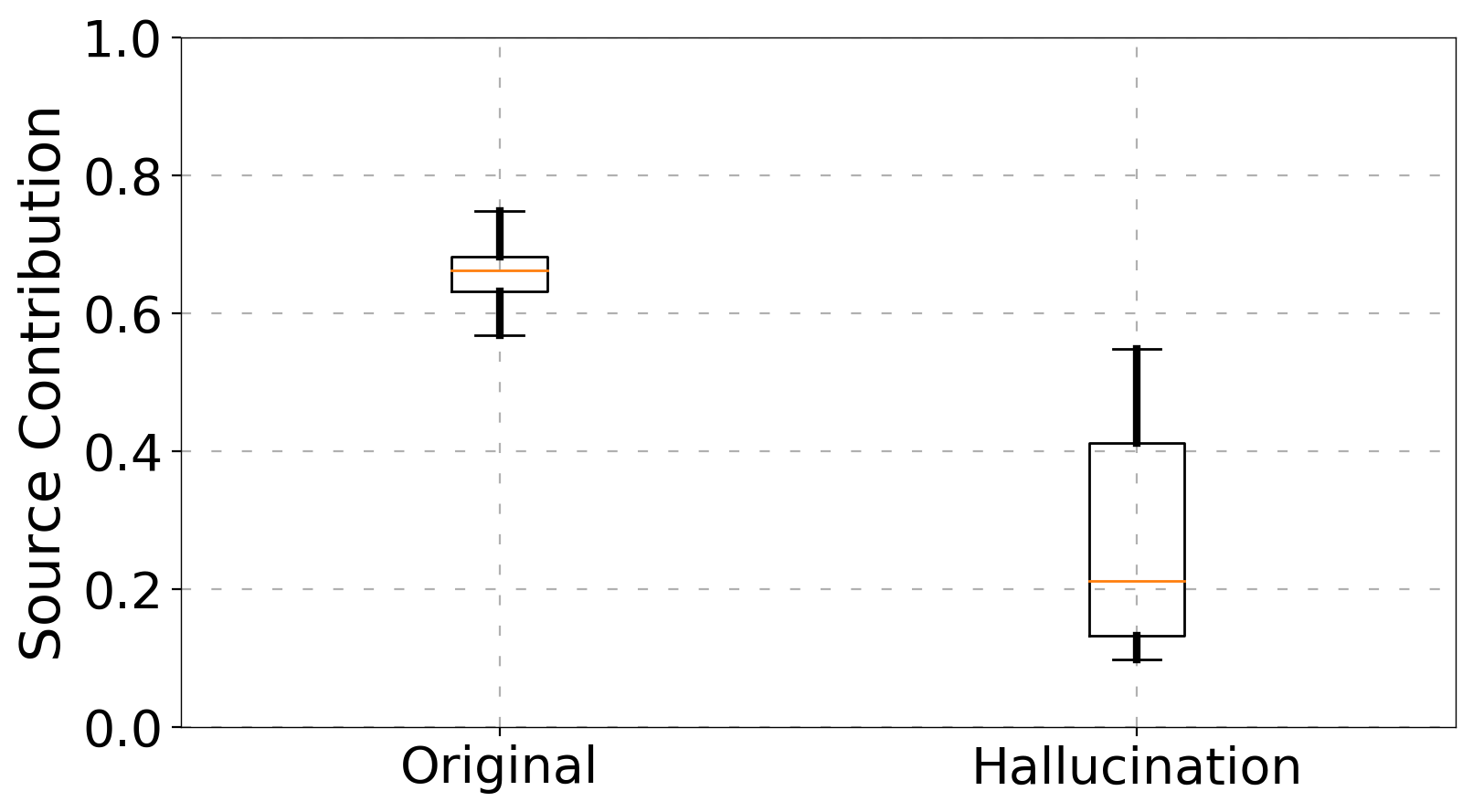}
\caption{Source contribution in sentences without and with induced perturbation in the bilingual model.}
\label{fig:perturbed_boxplot}
\end{center}
\end{figure}

\subsection{Analyzing Hallucinations}
\label{sec:hallucinations}

A common issue of NMT models is hallucination, which are translations that are disconnected from the source text, despite being fluent in the target language \cite{muller-etal-2020-domain}. Hallucinations should be reflected in our method as a drop in the contribution of the source sentence. Thus, in this section, we induce hallucination and measure the source sentence contribution with ALTI+.

\begin{figure*}[!t]
\begin{center}
\includegraphics[width=0.8\textwidth]{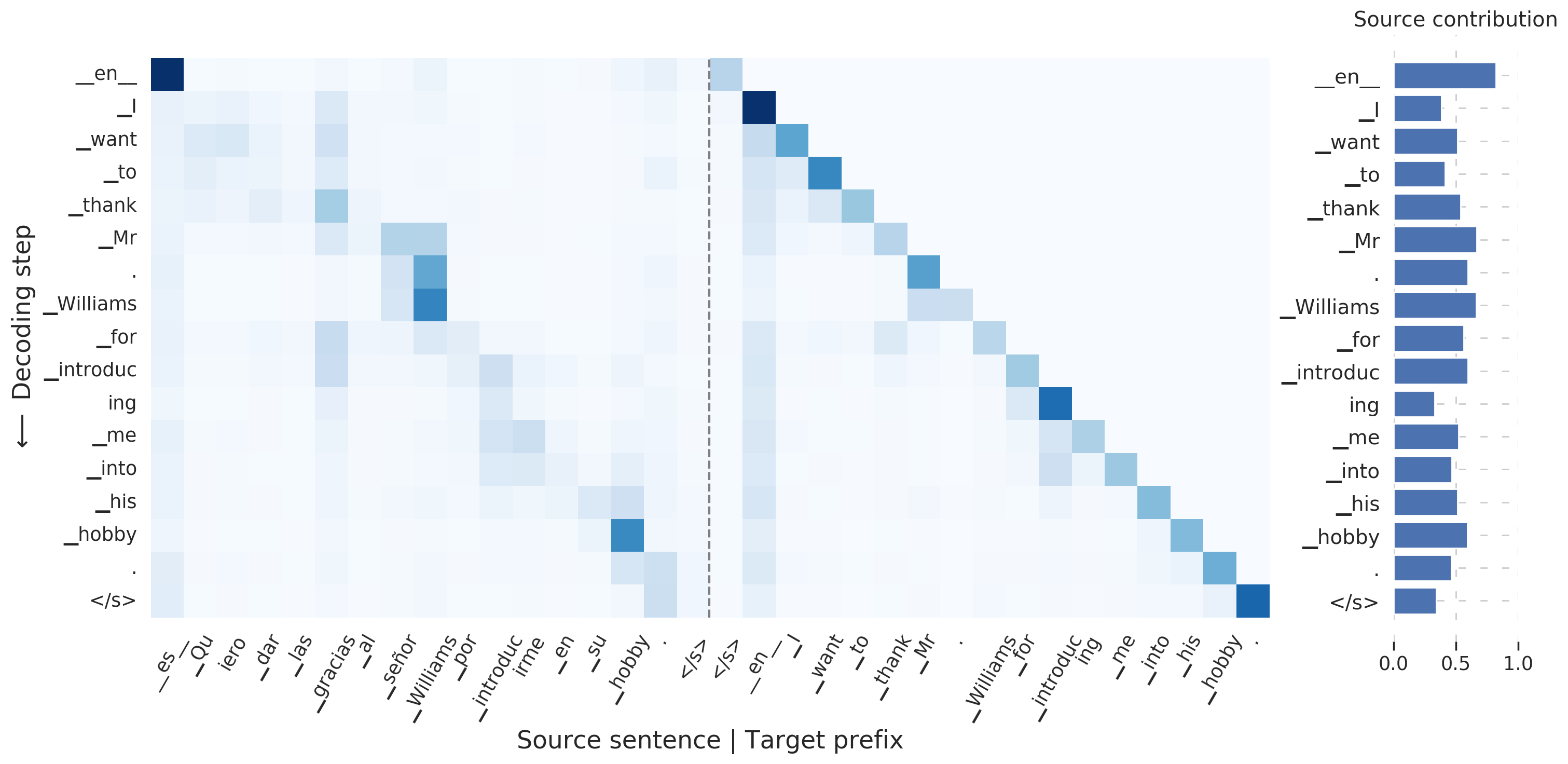}
\caption{ALTI+ for a Es-En example in the multilingual model.}
\label{fig:alti_nmt_multilingual}
\end{center}
\end{figure*}

To induce hallucination, we perturb the target prefix sequence of the bilingual model by adding the \texttt{<\text{unk}>} token. Then, we follow the algorithm proposed by \citet{Lee2018HallucinationsIN} to detect which perturbed translations are hallucinations. They measure BLEU score of the generated translation with and without perturbation. They fix a minimum threshold BLEU score for the original translations (20 BLEU in our experiments), and a maximum score for the perturbed translations (3 BLEU in our experiments). The model is considered to hallucinate when both translations satisfy the thresholds.

Analyzing ALTI+ contributions, we can confirm that the bilingual model largely ignores source tokens during hallucinations (\Cref{fig:hallucination_example,fig:perturbed_boxplot}).



%

\subsection{Multilingual Model Analysis}\label{sec:multilingual}

We analyze the behaviour of the multilingual model in different language pairs of \textsc{Flores-101} dataset. We include in the analysis high-resource languages, English (En), Spanish (Es), and French (Fr) and low-resource languages, Zulu (Zu) and Xhosa (Xh). High-resource languages have been defined in \cite{goyal-etal-2022-flores} as languages with available bi-text data beyond 100M samples, and low-resource languages are those with less than 1M.

\begin{figure}[!t]
\begin{center}
\includegraphics[width=0.48\textwidth]{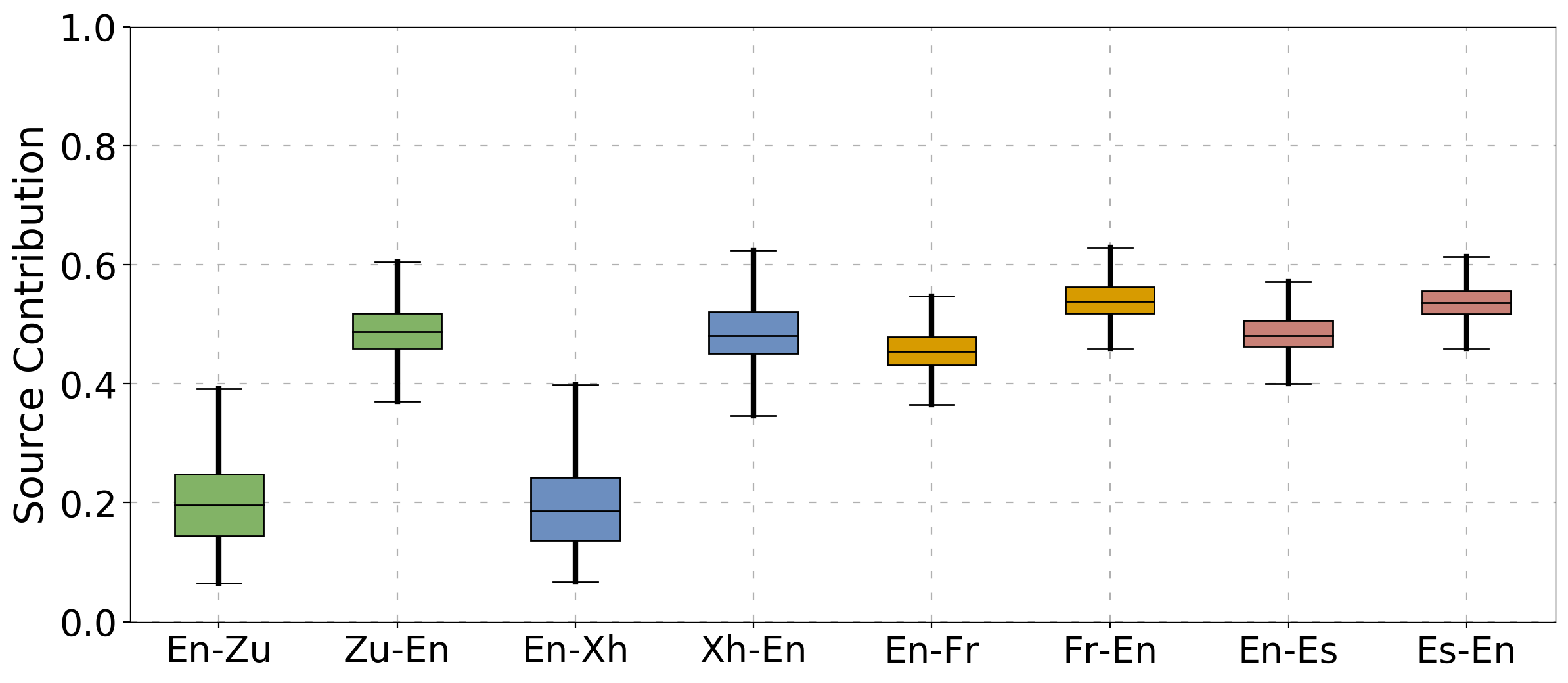}
\caption{Source sentence contribution in different language directions from the \textsc{Flores-101} devtest split.}
\label{fig:src_contrib_flores}
\end{center}
\end{figure}

\Cref{fig:alti_nmt_multilingual} shows an Es-En example in the multilingual model. We observe an almost uniform contribution of the language tags across different outputs. The only drop in its contribution seems to happen when translating proper nouns (e.g., "Mr. Williams") or anglicisms (e.g., "hobby"), which is observed for other language pairs too (\Cref{sec:appx_examples}), and repeated across the dataset. We hypothesize that the model doesn't need to rely on the language tag since these words appear across different languages. Dependencies between generated tokens are also observed, the prediction "for" relies on "thanks", "Williams" on "Mr." and "into" on "\underline{introduc}ing". The same example can be found in \Cref{sec:appx_examples} for En-Zu and Zu-En pairs.





Figure \ref{fig:src_contrib_flores} shows results of the source sentence contribution for En-Zu, En-Xh, En-Fr and En-Es pairs. We observe similar source contribution patterns between the high-resource pairs, and between those pairs involving a low-resource language. However, in the low-resource scenario, the source contribution is remarkably lower when translating from English. We hypothesize that, when the low-resource language is in the target prefix, the model tends to behave similarly to when it hallucinates (\Cref{fig:perturbed_boxplot}), ignoring the source. But, when a high-resource language (En) is in the target prefix, it is less likely to lose track of the source and, thus, less prone to enter hallucination mode. Low-resource language sentences in the target side may be seen by the model as target prefix perturbations (\Cref{sec:hallucinations}), although further research is required.

\section{Conclusions}
We propose ALTI+, an interpretability method for the encoder-decoder Transformer that provides token influences to the model predictions for the two input contexts: source sentence and target prefix. By applying ALTI+ to a bilingual and a multilingual NMT model we are able to discover insights into the behaviour of these black-box models. Unlike previous methods, we can now observe dependencies between tokens in the predicted sentence, and quantify the total contribution of each of the contexts. This allows a deeper exploration of current NMT models. Our findings include: the role of the source EOS ($\eos$) token as a mean to avoid incorporating source information, the absence of source contribution when producing hallucinations, and the lack of source contributions when translating from English to a low-resource language. ALTI+ overcomes the limitations of previous interpretability methods in NMT, and we believe it can help researchers and practitioners to better understand any encoder-decoder Transformer model.

\section*{Limitations}
ALTI+ is able to measure the amount of contextual information in each layer representation of the Transformer. We use the influences of each input token to the last layer representation for evaluating input attributions for the model prediction. However, our method does not consider the softmax layer on top of the Transformer. Therefore, ALTI+ doesn't provide explanations for each of the output classes (target vocabulary), as opposed to gradient-based methods.
\section*{Ethical Considerations}
ALTI+ provides explanations about input attributions in the Encoder-Decoder Transformer. By itself, we are not aware of any ethical implications of the methodology, which does not take into account any subjective priors. We perform experiments in Machine Translation. While we do not study biases in this application, we know they exist \cite{AAAI_bias}. In the future, we plan to further explore and mitigate them by using the information of source input attributions that ALTI+ provides. Also, understanding hallucinations by means of ALTI+ can help to avoid catastrophic and unsafe translations.

\section{Acknowledgements}
We would like to thank the anonymous reviewers for their useful comments. Javier Ferrando and Gerard I. Gállego are supported by the Spanish Ministerio de Ciencia e Innovación through the project PID2019-107579RB-I00 / AEI / 10.13039/501100011033.
\bibliography{anthology,custom}

\begin{thebibliography}{35}
\expandafter\ifx\csname natexlab\endcsname\relax\def\natexlab#1{#1}\fi

\bibitem[{Abnar and Zuidema(2020)}]{abnar-zuidema-2020-quantifying}
Samira Abnar and Willem Zuidema. 2020.
\newblock \href {https://doi.org/10.18653/v1/2020.acl-main.385} {Quantifying
  attention flow in transformers}.
\newblock In \emph{Proceedings of the 58th Annual Meeting of the Association
  for Computational Linguistics}, pages 4190--4197, Online. Association for
  Computational Linguistics.

\bibitem[{Bach et~al.(2015)Bach, Binder, Montavon, Klauschen, Müller, and
  Samek}]{LRP_bach}
Sebastian Bach, Alexander Binder, Grégoire Montavon, Frederick Klauschen,
  Klaus-Robert Müller, and Wojciech Samek. 2015.
\newblock \href {https://doi.org/10.1371/journal.pone.0130140} {On pixel-wise
  explanations for non-linear classifier decisions by layer-wise relevance
  propagation}.
\newblock \emph{PLOS ONE}, 10(7):1--46.

\bibitem[{Brown et~al.(2020)Brown, Mann, Ryder, Subbiah, Kaplan, Dhariwal,
  Neelakantan, Shyam, Sastry, Askell, Agarwal, Herbert-Voss, Krueger, Henighan,
  Child, Ramesh, Ziegler, Wu, Winter, Hesse, Chen, Sigler, Litwin, Gray, Chess,
  Clark, Berner, McCandlish, Radford, Sutskever, and
  Amodei}]{NEURIPS2020_1457c0d6}
Tom Brown, Benjamin Mann, Nick Ryder, Melanie Subbiah, Jared~D Kaplan, Prafulla
  Dhariwal, Arvind Neelakantan, Pranav Shyam, Girish Sastry, Amanda Askell,
  Sandhini Agarwal, Ariel Herbert-Voss, Gretchen Krueger, Tom Henighan, Rewon
  Child, Aditya Ramesh, Daniel Ziegler, Jeffrey Wu, Clemens Winter, Chris
  Hesse, Mark Chen, Eric Sigler, Mateusz Litwin, Scott Gray, Benjamin Chess,
  Jack Clark, Christopher Berner, Sam McCandlish, Alec Radford, Ilya Sutskever,
  and Dario Amodei. 2020.
\newblock \href
  {https://proceedings.neurips.cc/paper/2020/file/1457c0d6bfcb4967418bfb8ac142f64a-Paper.pdf}
  {Language models are few-shot learners}.
\newblock In \emph{Advances in Neural Information Processing Systems},
  volume~33, pages 1877--1901. Curran Associates, Inc.

\bibitem[{Brunner et~al.(2020)Brunner, Liu, Pascual, Richter, Ciaramita, and
  Wattenhofer}]{Brunner2020On}
Gino Brunner, Yang Liu, Damian Pascual, Oliver Richter, Massimiliano Ciaramita,
  and Roger Wattenhofer. 2020.
\newblock \href {https://openreview.net/forum?id=BJg1f6EFDB} {On
  identifiability in transformers}.
\newblock In \emph{International Conference on Learning Representations}.

\bibitem[{Chefer et~al.(2021{\natexlab{a}})Chefer, Gur, and
  Wolf}]{Chefer_2021_ICCV}
Hila Chefer, Shir Gur, and Lior Wolf. 2021{\natexlab{a}}.
\newblock \href
  {https://openaccess.thecvf.com/content/ICCV2021/papers/Chefer_Generic_Attention-Model_Explainability_for_Interpreting_Bi-Modal_and_Encoder-Decoder_Transformers_ICCV_2021_paper.pdf}
  {Generic attention-model explainability for interpreting bi-modal and
  encoder-decoder transformers}.
\newblock In \emph{Proceedings of the IEEE/CVF International Conference on
  Computer Vision (ICCV)}, pages 397--406.

\bibitem[{Chefer et~al.(2021{\natexlab{b}})Chefer, Gur, and
  Wolf}]{Chefer_2021_CVPR}
Hila Chefer, Shir Gur, and Lior Wolf. 2021{\natexlab{b}}.
\newblock \href
  {https://openaccess.thecvf.com/content/CVPR2021/papers/Chefer_Transformer_Interpretability_Beyond_Attention_Visualization_CVPR_2021_paper.pdf}
  {Transformer interpretability beyond attention visualization}.
\newblock In \emph{Proceedings of the IEEE/CVF Conference on Computer Vision
  and Pattern Recognition (CVPR)}, pages 782--791.

\bibitem[{Chen et~al.(2020)Chen, Liu, Chen, Jiang, and
  Liu}]{chen-etal-2020-accurate}
Yun Chen, Yang Liu, Guanhua Chen, Xin Jiang, and Qun Liu. 2020.
\newblock \href {https://doi.org/10.18653/v1/2020.emnlp-main.42} {Accurate word
  alignment induction from neural machine translation}.
\newblock In \emph{Proceedings of the 2020 Conference on Empirical Methods in
  Natural Language Processing (EMNLP)}, pages 566--576, Online. Association for
  Computational Linguistics.

\bibitem[{Clark et~al.(2019)Clark, Khandelwal, Levy, and
  Manning}]{clark-etal-2019-bert}
Kevin Clark, Urvashi Khandelwal, Omer Levy, and Christopher~D. Manning. 2019.
\newblock \href {https://doi.org/10.18653/v1/W19-4828} {What does {BERT} look
  at? an analysis of {BERT}{'}s attention}.
\newblock In \emph{Proceedings of the 2019 ACL Workshop BlackboxNLP: Analyzing
  and Interpreting Neural Networks for NLP}, pages 276--286, Florence, Italy.
  Association for Computational Linguistics.

\bibitem[{Costa-jussà et~al.(2022)Costa-jussà, Escolano, Basta, Ferrando,
  Batlle, and Kharitonova}]{AAAI_bias}
Marta~R. Costa-jussà, Carlos Escolano, Christine Basta, Javier Ferrando, Roser
  Batlle, and Ksenia Kharitonova. 2022.
\newblock \href {https://doi.org/10.1609/aaai.v36i11.21442} {Interpreting
  gender bias in neural machine translation: Multilingual architecture
  matters}.
\newblock \emph{Proceedings of the AAAI Conference on Artificial Intelligence},
  36(11):11855--11863.

\bibitem[{Devlin et~al.(2019)Devlin, Chang, Lee, and
  Toutanova}]{devlin-etal-2019-bert}
Jacob Devlin, Ming-Wei Chang, Kenton Lee, and Kristina Toutanova. 2019.
\newblock \href {https://doi.org/10.18653/v1/N19-1423} {{BERT}: Pre-training of
  deep bidirectional transformers for language understanding}.
\newblock In \emph{Proceedings of the 2019 Conference of the North {A}merican
  Chapter of the Association for Computational Linguistics: Human Language
  Technologies, Volume 1 (Long and Short Papers)}, pages 4171--4186,
  Minneapolis, Minnesota. Association for Computational Linguistics.

\bibitem[{Ding et~al.(2019)Ding, Xu, and Koehn}]{ding-etal-2019-saliency}
Shuoyang Ding, Hainan Xu, and Philipp Koehn. 2019.
\newblock \href {https://doi.org/10.18653/v1/W19-5201} {Saliency-driven word
  alignment interpretation for neural machine translation}.
\newblock In \emph{Proceedings of the Fourth Conference on Machine Translation
  (Volume 1: Research Papers)}, pages 1--12, Florence, Italy. Association for
  Computational Linguistics.

\bibitem[{Fan et~al.(2021)Fan, Bhosale, Schwenk, Ma, El-Kishky, Goyal, Baines,
  Celebi, Wenzek, Chaudhary, Goyal, Birch, Liptchinsky, Edunov, Auli, and
  Joulin}]{m2m_100}
Angela Fan, Shruti Bhosale, Holger Schwenk, Zhiyi Ma, Ahmed El-Kishky,
  Siddharth Goyal, Mandeep Baines, Onur Celebi, Guillaume Wenzek, Vishrav
  Chaudhary, Naman Goyal, Tom Birch, Vitaliy Liptchinsky, Sergey Edunov,
  Michael Auli, and Armand Joulin. 2021.
\newblock \href {http://jmlr.org/papers/v22/20-1307.html} {Beyond
  english-centric multilingual machine translation}.
\newblock \emph{Journal of Machine Learning Research}, 22(107):1--48.

\bibitem[{Ferrando and
  Costa-juss{\`a}(2021)}]{ferrando-costa-jussa-2021-attention-weights}
Javier Ferrando and Marta~R. Costa-juss{\`a}. 2021.
\newblock \href {https://doi.org/10.18653/v1/2021.findings-emnlp.39} {Attention
  weights in transformer {NMT} fail aligning words between sequences but
  largely explain model predictions}.
\newblock In \emph{Findings of the Association for Computational Linguistics:
  EMNLP 2021}, pages 434--443, Punta Cana, Dominican Republic. Association for
  Computational Linguistics.

\bibitem[{Ferrando et~al.(2022)Ferrando, Gállego, and
  Costa-jussà}]{ferrando2022measuring}
Javier Ferrando, Gerard~I. Gállego, and Marta~R. Costa-jussà. 2022.
\newblock \href {https://doi.org/10.48550/ARXIV.2203.04212} {Measuring the
  mixing of contextual information in the transformer}.

\bibitem[{Garg et~al.(2019)Garg, Peitz, Nallasamy, and
  Paulik}]{garg-etal-2019-jointly}
Sarthak Garg, Stephan Peitz, Udhyakumar Nallasamy, and Matthias Paulik. 2019.
\newblock \href {https://doi.org/10.18653/v1/D19-1453} {Jointly learning to
  align and translate with transformer models}.
\newblock In \emph{Proceedings of the 2019 Conference on Empirical Methods in
  Natural Language Processing and the 9th International Joint Conference on
  Natural Language Processing (EMNLP-IJCNLP)}, pages 4453--4462, Hong Kong,
  China. Association for Computational Linguistics.

\bibitem[{Goyal et~al.(2022)Goyal, Gao, Chaudhary, Chen, Wenzek, Ju, Krishnan,
  Ranzato, Guzm{\'a}n, and Fan}]{goyal-etal-2022-flores}
Naman Goyal, Cynthia Gao, Vishrav Chaudhary, Peng-Jen Chen, Guillaume Wenzek,
  Da~Ju, Sanjana Krishnan, Marc{'}Aurelio Ranzato, Francisco Guzm{\'a}n, and
  Angela Fan. 2022.
\newblock \href {https://doi.org/10.1162/tacl_a_00474} {The {F}lores-101
  evaluation benchmark for low-resource and multilingual machine translation}.
\newblock \emph{Transactions of the Association for Computational Linguistics},
  10:522--538.

\bibitem[{Jain and Wallace(2019)}]{jain-wallace-2019-attention}
Sarthak Jain and Byron~C. Wallace. 2019.
\newblock \href {https://doi.org/10.18653/v1/N19-1357} {{A}ttention is not
  {E}xplanation}.
\newblock In \emph{Proceedings of the 2019 Conference of the North {A}merican
  Chapter of the Association for Computational Linguistics: Human Language
  Technologies, Volume 1 (Long and Short Papers)}, pages 3543--3556,
  Minneapolis, Minnesota. Association for Computational Linguistics.

\bibitem[{Kobayashi et~al.(2020)Kobayashi, Kuribayashi, Yokoi, and
  Inui}]{kobayashi-etal-2020-attention}
Goro Kobayashi, Tatsuki Kuribayashi, Sho Yokoi, and Kentaro Inui. 2020.
\newblock \href {https://doi.org/10.18653/v1/2020.emnlp-main.574} {Attention is
  not only a weight: Analyzing transformers with vector norms}.
\newblock In \emph{Proceedings of the 2020 Conference on Empirical Methods in
  Natural Language Processing (EMNLP)}, pages 7057--7075, Online. Association
  for Computational Linguistics.

\bibitem[{Kobayashi et~al.(2021)Kobayashi, Kuribayashi, Yokoi, and
  Inui}]{kobayashi-etal-2021-incorporating}
Goro Kobayashi, Tatsuki Kuribayashi, Sho Yokoi, and Kentaro Inui. 2021.
\newblock \href {https://doi.org/10.18653/v1/2021.emnlp-main.373}
  {{I}ncorporating {R}esidual and {N}ormalization {L}ayers into {A}nalysis of
  {M}asked {L}anguage {M}odels}.
\newblock In \emph{Proceedings of the 2021 Conference on Empirical Methods in
  Natural Language Processing}, pages 4547--4568, Online and Punta Cana,
  Dominican Republic. Association for Computational Linguistics.

\bibitem[{Lee et~al.(2018)Lee, Firat, Agarwal, Fannjiang, and
  Sussillo}]{Lee2018HallucinationsIN}
Katherine Lee, Orhan Firat, Ashish Agarwal, Clara Fannjiang, and David
  Sussillo. 2018.
\newblock \href {https://openreview.net/pdf?id=SJxTk3vB3m} {Hallucinations in
  neural machine translation}.
\newblock \emph{NIPS 2018 Interpretability and Robustness for Audio, Speech and
  Language Workshop}.

\bibitem[{Li et~al.(2019)Li, Li, Liu, Meng, and Shi}]{li-etal-2019-word}
Xintong Li, Guanlin Li, Lemao Liu, Max Meng, and Shuming Shi. 2019.
\newblock \href {https://doi.org/10.18653/v1/P19-1124} {On the word alignment
  from neural machine translation}.
\newblock In \emph{Proceedings of the 57th Annual Meeting of the Association
  for Computational Linguistics}, pages 1293--1303, Florence, Italy.
  Association for Computational Linguistics.

\bibitem[{Liu et~al.(2019)Liu, Ott, Goyal, Du, Joshi, Chen, Levy, Lewis,
  Zettlemoyer, and Stoyanov}]{DBLP:journals/corr/abs-1907-11692}
Yinhan Liu, Myle Ott, Naman Goyal, Jingfei Du, Mandar Joshi, Danqi Chen, Omer
  Levy, Mike Lewis, Luke Zettlemoyer, and Veselin Stoyanov. 2019.
\newblock \href {http://arxiv.org/abs/1907.11692} {Roberta: {A} robustly
  optimized {BERT} pretraining approach}.
\newblock \emph{CoRR}, abs/1907.11692.

\bibitem[{M{\"u}ller et~al.(2020)M{\"u}ller, Rios, and
  Sennrich}]{muller-etal-2020-domain}
Mathias M{\"u}ller, Annette Rios, and Rico Sennrich. 2020.
\newblock \href {https://aclanthology.org/2020.amta-research.14} {Domain
  robustness in neural machine translation}.
\newblock In \emph{Proceedings of the 14th Conference of the Association for
  Machine Translation in the Americas (Volume 1: Research Track)}, pages
  151--164, Virtual. Association for Machine Translation in the Americas.

\bibitem[{Ott et~al.(2019)Ott, Edunov, Baevski, Fan, Gross, Ng, Grangier, and
  Auli}]{ott-etal-2019-fairseq}
Myle Ott, Sergey Edunov, Alexei Baevski, Angela Fan, Sam Gross, Nathan Ng,
  David Grangier, and Michael Auli. 2019.
\newblock \href {https://doi.org/10.18653/v1/N19-4009} {fairseq: A fast,
  extensible toolkit for sequence modeling}.
\newblock In \emph{Proceedings of the 2019 Conference of the North {A}merican
  Chapter of the Association for Computational Linguistics (Demonstrations)},
  pages 48--53, Minneapolis, Minnesota. Association for Computational
  Linguistics.

\bibitem[{Raffel et~al.(2020)Raffel, Shazeer, Roberts, Lee, Narang, Matena,
  Zhou, Li, and Liu}]{T5_raffel}
Colin Raffel, Noam Shazeer, Adam Roberts, Katherine Lee, Sharan Narang, Michael
  Matena, Yanqi Zhou, Wei Li, and Peter~J. Liu. 2020.
\newblock \href {http://jmlr.org/papers/v21/20-074.html} {Exploring the limits
  of transfer learning with a unified text-to-text transformer}.
\newblock \emph{Journal of Machine Learning Research}, 21(140):1--67.

\bibitem[{Raganato and Tiedemann(2018)}]{raganato-tiedemann-2018-analysis}
Alessandro Raganato and J{\"o}rg Tiedemann. 2018.
\newblock \href {https://doi.org/10.18653/v1/W18-5431} {An analysis of encoder
  representations in transformer-based machine translation}.
\newblock In \emph{Proceedings of the 2018 {EMNLP} Workshop {B}lackbox{NLP}:
  Analyzing and Interpreting Neural Networks for {NLP}}, pages 287--297,
  Brussels, Belgium. Association for Computational Linguistics.

\bibitem[{Sennrich et~al.(2016)Sennrich, Haddow, and
  Birch}]{sennrich-etal-2016-neural}
Rico Sennrich, Barry Haddow, and Alexandra Birch. 2016.
\newblock \href {https://doi.org/10.18653/v1/P16-1162} {Neural machine
  translation of rare words with subword units}.
\newblock In \emph{Proceedings of the 54th Annual Meeting of the Association
  for Computational Linguistics (Volume 1: Long Papers)}, pages 1715--1725,
  Berlin, Germany. Association for Computational Linguistics.

\bibitem[{Serrano and Smith(2019)}]{serrano-smith-2019-attention}
Sofia Serrano and Noah~A. Smith. 2019.
\newblock \href {https://doi.org/10.18653/v1/P19-1282} {Is attention
  interpretable?}
\newblock In \emph{Proceedings of the 57th Annual Meeting of the Association
  for Computational Linguistics}, pages 2931--2951, Florence, Italy.
  Association for Computational Linguistics.

\bibitem[{Vaswani et~al.(2017)Vaswani, Shazeer, Parmar, Uszkoreit, Jones,
  Gomez, Kaiser, and Polosukhin}]{NIPS2017_3f5ee243}
Ashish Vaswani, Noam Shazeer, Niki Parmar, Jakob Uszkoreit, Llion Jones,
  Aidan~N Gomez, \L~ukasz Kaiser, and Illia Polosukhin. 2017.
\newblock \href
  {https://proceedings.neurips.cc/paper/2017/file/3f5ee243547dee91fbd053c1c4a845aa-Paper.pdf}
  {Attention is all you need}.
\newblock In \emph{Advances in Neural Information Processing Systems},
  volume~30. Curran Associates, Inc.

\bibitem[{Vilar et~al.(2006)Vilar, Popovic, and Ney}]{Vilar2006AERDW}
David Vilar, Maja Popovic, and Hermann Ney. 2006.
\newblock \href {https://aclanthology.org/2006.iwslt-papers.7} {{AER}: do we
  need to {``}improve{''} our alignments?}
\newblock In \emph{Proceedings of the Third International Workshop on Spoken
  Language Translation: Papers}, Kyoto, Japan.

\bibitem[{Voita et~al.(2019)Voita, Sennrich, and
  Titov}]{voita-etal-2019-bottom}
Elena Voita, Rico Sennrich, and Ivan Titov. 2019.
\newblock \href {https://doi.org/10.18653/v1/D19-1448} {The bottom-up evolution
  of representations in the transformer: A study with machine translation and
  language modeling objectives}.
\newblock In \emph{Proceedings of the 2019 Conference on Empirical Methods in
  Natural Language Processing and the 9th International Joint Conference on
  Natural Language Processing (EMNLP-IJCNLP)}, pages 4396--4406, Hong Kong,
  China. Association for Computational Linguistics.

\bibitem[{Voita et~al.(2021{\natexlab{a}})Voita, Sennrich, and
  Titov}]{voita-etal-2021-analyzing}
Elena Voita, Rico Sennrich, and Ivan Titov. 2021{\natexlab{a}}.
\newblock \href {https://doi.org/10.18653/v1/2021.acl-long.91} {Analyzing the
  source and target contributions to predictions in neural machine
  translation}.
\newblock In \emph{Proceedings of the 59th Annual Meeting of the Association
  for Computational Linguistics and the 11th International Joint Conference on
  Natural Language Processing (Volume 1: Long Papers)}, pages 1126--1140,
  Online. Association for Computational Linguistics.

\bibitem[{Voita et~al.(2021{\natexlab{b}})Voita, Sennrich, and
  Titov}]{voita-etal-2021-language}
Elena Voita, Rico Sennrich, and Ivan Titov. 2021{\natexlab{b}}.
\newblock \href {https://doi.org/10.18653/v1/2021.emnlp-main.667} {Language
  modeling, lexical translation, reordering: The training process of {NMT}
  through the lens of classical {SMT}}.
\newblock In \emph{Proceedings of the 2021 Conference on Empirical Methods in
  Natural Language Processing}, pages 8478--8491, Online and Punta Cana,
  Dominican Republic. Association for Computational Linguistics.

\bibitem[{Voita et~al.(2018)Voita, Serdyukov, Sennrich, and
  Titov}]{voita-etal-2018-context}
Elena Voita, Pavel Serdyukov, Rico Sennrich, and Ivan Titov. 2018.
\newblock \href {https://doi.org/10.18653/v1/P18-1117} {Context-aware neural
  machine translation learns anaphora resolution}.
\newblock In \emph{Proceedings of the 56th Annual Meeting of the Association
  for Computational Linguistics (Volume 1: Long Papers)}, pages 1264--1274,
  Melbourne, Australia. Association for Computational Linguistics.

\bibitem[{Zenkel et~al.(2019)Zenkel, Wuebker, and DeNero}]{Zenkel_2019}
Thomas Zenkel, Joern Wuebker, and John DeNero. 2019.
\newblock \href {http://arxiv.org/abs/1901.11359} {Adding interpretable
  attention to neural translation models improves word alignment}.
\newblock \emph{CoRR}, abs/1901.11359.

\end{thebibliography}
\bibliographystyle{acl_natbib}

\clearpage

\appendix


\section{ALTI}\label{apx:alti}
\subsection{Layer Normalization}\label{apx:ln}
The Layer normalization operation over input $\bm x$ can be defined as: $\text{LN}(\bm x)=\frac{\bm x-\mu(\bm x)}{\sigma(\bm x)} \odot \mathbf{\gamma}+ \mathbf{\beta}$, where $\mu$ computes the mean, $\sigma$ the standard deviation, and $\bm \gamma$ and $\bm \beta$ refer to an element-wise transformation and bias respectively.
$\text{LN}(\bm x)$ can be decomposed into $\frac{1}{\sigma(\bm x)}\mathbf{L} \bm x + \bm \beta$, where $\mathbf{L}$ is a linear transformation including the mean and element wise multiplication.

Given a sum of vectors $\sum_j \bm x_j$ as input to LN we can rewrite the expression as:
\begin{align*}
    \text{LN}(\sum_j \bm x_j) &= \frac{1}{\sigma(\sum_j \bm x_j)}\mathbf{L} \sum_j \bm x_j + \bm \beta\\
    &= \sum_j \frac{1}{\sigma(\sum_j \bm x_j)} \mathbf{L}\bm x_j + \bm \beta\\
    &= \sum_j L(\bm x_j) + \bm \beta
\end{align*}

\subsection{Full derivation}\label{apx:full_derivation}
\begin{align*}
\widetilde{\bm x}_i &= \text{LN}\Bigg(\sum_{j=1}^J \sum^H_{h=1} \mathbf{W}_O^{h} \alpha_{i,j}^{h} \mathbf{W}_V^{h}\bm x_{j} + \bm b_O + \bm x_{i}\Bigg)\\
&= \text{LN}\Bigg(\sum_{j=1}^J F_{i}(\bm x_j) + \bm b_O + \bm x_{i}\Bigg)\\
&= \sum_{j=1}^J L(F_{i}(\bm x_j)) + L(\bm b_O) + L(\bm x_{i}) + \bm \beta
\end{align*}

Defining $\bm \epsilon = L(\bm b_O) + \bm \beta$ we get to the expression in \Cref{eq:post_layer_transformed_vectors}:
\begin{equation}
\widetilde{\bm x}_i = \sum_{j=1}^J T_i(\bm x_j) + \bm \epsilon
\end{equation}

\newpage

\section{Values Norms}
\label{sec:values_norms}
\begin{figure}[!h]
\begin{center}
\includegraphics[width=0.48\textwidth]{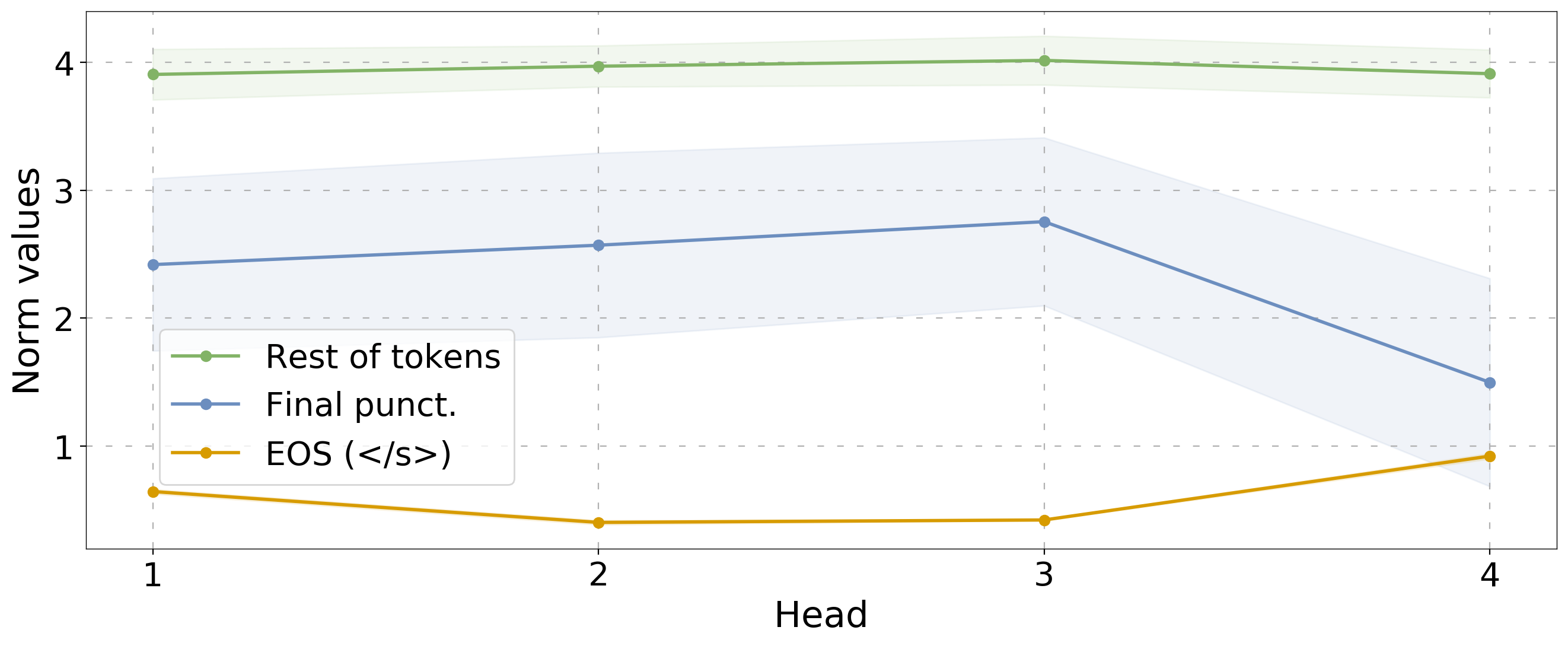}
\caption{Norm of the value vectors (from encoder outputs) in the cross-attention of the alignment layer. We provide mean and SD for each head in the bilingual model. Similar patterns are observed across layers, and in the multilingual model.}
\label{fig:norm_values}
\end{center}
\end{figure}

\section{Examples}
\label{sec:appx_examples}

\begin{figure*}[h!]
\begin{center}
\includegraphics[width=0.98\textwidth]{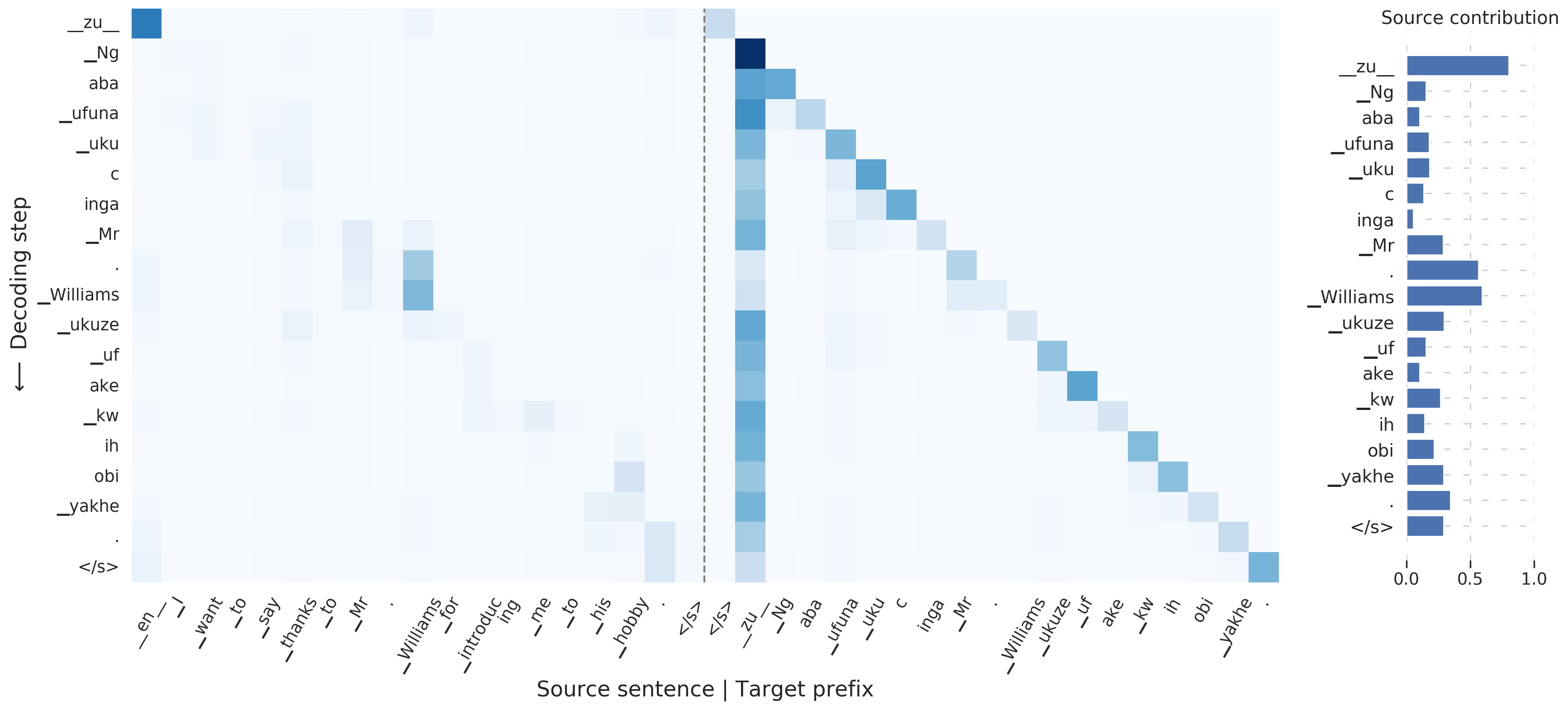}
\caption{ALTI+ for a En-Zu example in the multilingual model.}
\label{fig:alti_enzu}
\end{center}
\end{figure*}

\begin{figure*}[h!]
\begin{center}
\includegraphics[width=0.98\textwidth]{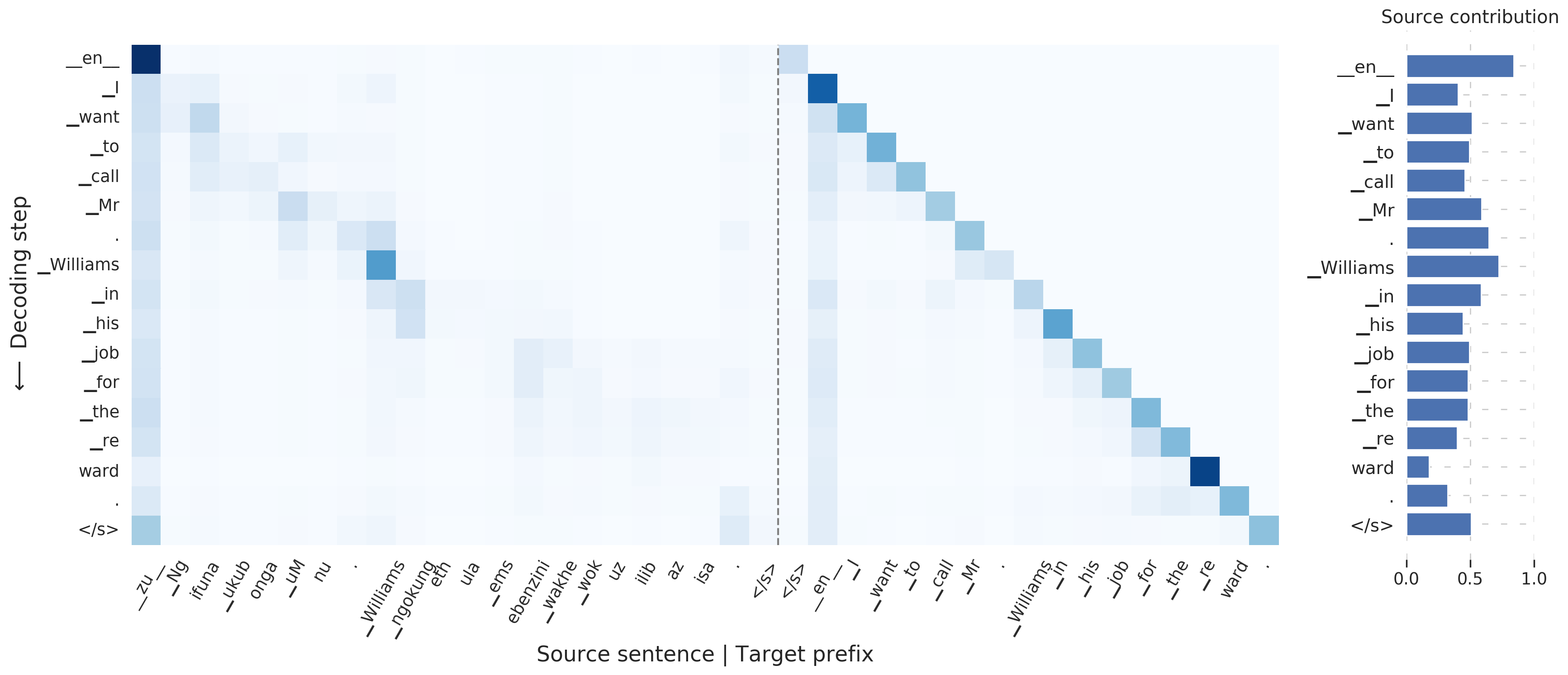}
\caption{ALTI+ for a Zu-En example in the multilingual model.}
\label{fig:alti_zuen}
\end{center}
\end{figure*}

\begin{figure*}[h!]
\begin{center}
\includegraphics[width=0.98\textwidth]{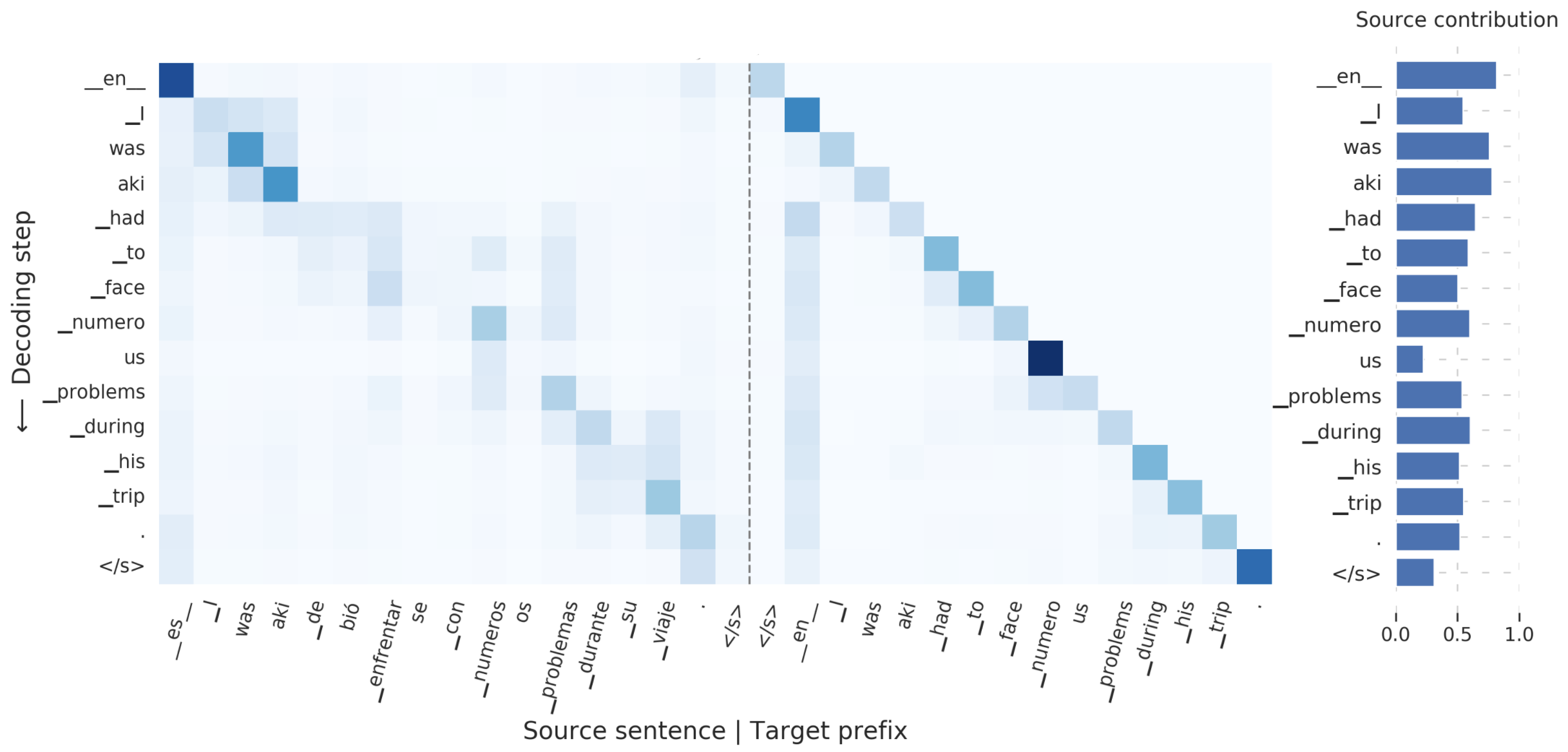}
\caption{ALTI+ for a Es-En example in the multilingual model.}
\label{fig:alti_es_en_Iwasaki}
\end{center}
\end{figure*}

\begin{figure*}[h!]
\begin{center}
\includegraphics[width=0.98\textwidth]{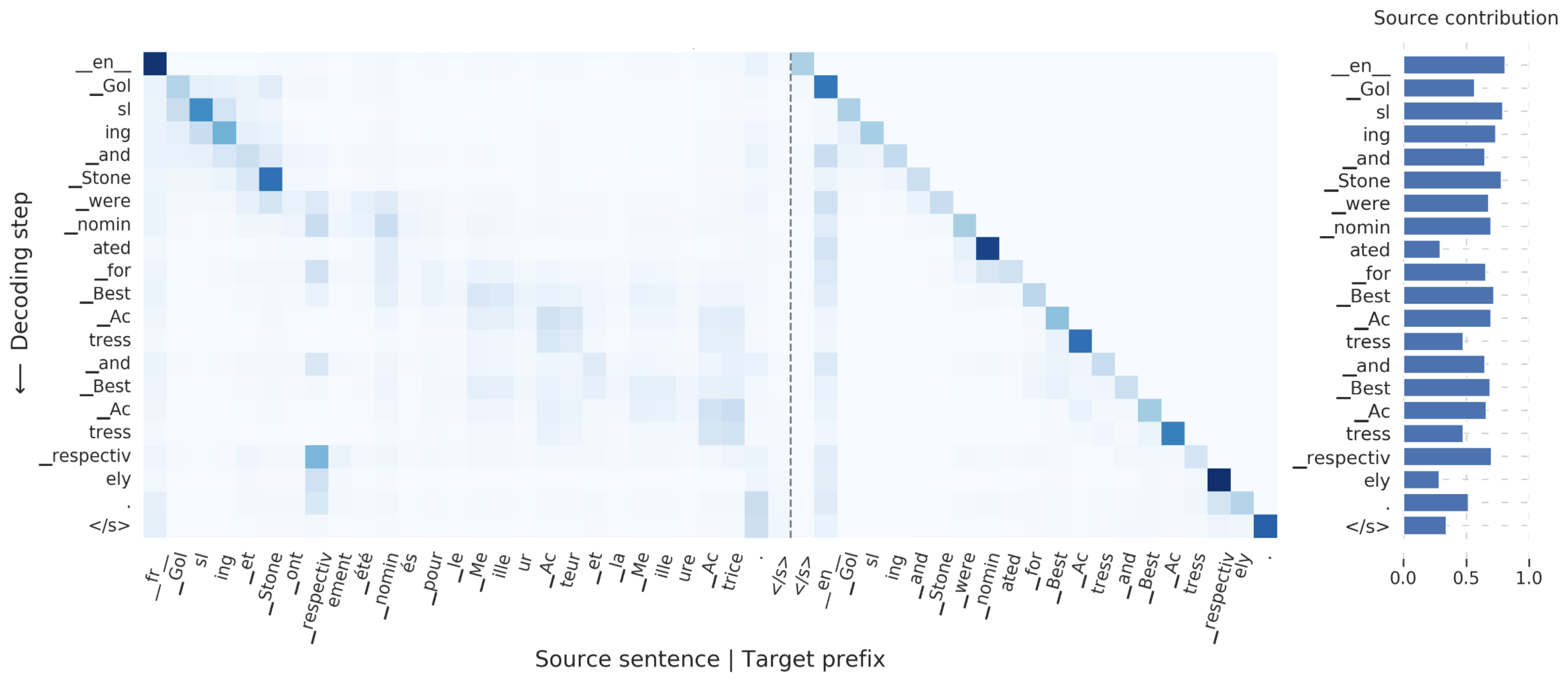}
\caption{ALTI+ for a Fr-En example in the multilingual model.}
\label{fig:alti_fr_en}
\end{center}
\end{figure*}

We include examples for the En-Zu language pair in the multilingual model in \Cref{fig:alti_enzu} and \ref{fig:alti_zuen}, as well as for Es-En in \Cref{fig:alti_es_en_Iwasaki} and Fr-En in \Cref{fig:alti_fr_en}.

\end{document}